%% file: main.tex
\documentclass{article}

\PassOptionsToPackage{numbers, compress}{natbib}

\usepackage[final]{neurips_data_2023}

\usepackage[utf8]{inputenc} 
\usepackage[T1]{fontenc}    
\usepackage{hyperref}       
\usepackage{url}            
\usepackage{booktabs}       
\usepackage{amsfonts}       
\usepackage{nicefrac}       
\usepackage{microtype}      
\usepackage{xcolor}         
\usepackage{graphicx}       
\usepackage{tikz}
\usepackage[many]{tcolorbox}
\usepackage{textcomp}
\usepackage{multirow}
\usepackage{wrapfig}
\usepackage{caption}
\usepackage{soul}

\definecolor{CB_gray}{gray}{0.5}

\tcbset{prompt/.style={
    enhanced,
    size=fbox,
    boxrule=3pt,
    arc=2mm,
    auto outer arc,
    left=10pt,
    right=10pt,
    top=10pt,
    bottom=10pt,
    fontupper=\fontfamily{cmtt}\selectfont, 
    colback=CB_gray!10,
    colframe=CB_gray!25,
    coltitle=CB_gray!100, 
}}

\tcbset{vignette/.style={
    enhanced,
    size=fbox,
    boxrule=3pt,
    arc=2mm,
    auto outer arc,
    left=10pt,
    right=10pt,
    top=10pt,
    bottom=10pt,
    fontupper=\fontfamily{Avenir}\selectfont, 
    colback=CB_gray!10,
    colframe=CB_gray!25,
    coltitle=CB_gray!100, 
}}

\def\Snospace~{\S{}}

\definecolor{truecolor}{RGB}{31, 120, 180}
\definecolor{falsecolor}{RGB}{213, 94, 0}
\definecolor{fbonly}{RGB}{120, 120, 120}
\definecolor{desire}{RGB}{118, 165, 175}
\definecolor{belief}{RGB}{224, 102, 102}
\definecolor{action}{RGB}{106, 168, 79}
\definecolor{percept}{RGB}{180, 167, 214}

\definecolor{coco1}{HTML}{D9E4EC}
\definecolor{coco2}{HTML}{B7CFDC}
\definecolor{coco3}{HTML}{6AABD2}
\definecolor{coco4}{HTML}{385E72}
\hypersetup{
    colorlinks=true,
    linkcolor=coco3,
    filecolor=coco4,      
    urlcolor=coco3,
    citecolor=coco3,
}

\title{Understanding Social Reasoning in Language Models with Language Models}
\author{%
  Kanishk Gandhi \thanks{Equal Contribution.}
\quad  J.-Philipp Fr\"anken \footnotemark[1]
\quad Tobias Gerstenberg
\quad Noah D. Goodman \\
Stanford University \\
\texttt{\{kanishk.gandhi, jphilipp\}@stanford.edu}
}

\begin{document}

\maketitle

\begin{abstract}
\input{sections/00_abstract}
\end{abstract}
\vspace{-6mm}
\section{Introduction}
\vspace{-2mm}
\label{sec:intro}
\input{sections/01_introduction}
\vspace{-2mm}
\section{Related Work}
\vspace{-2mm}
\label{sec:rel}
\input{sections/02_related-work}
\vspace{-2mm}
\section{Model-Written Evaluations with Causal Templates}
\vspace{-2mm}
\label{sec:methods}
\input{sections/03_method}

\vspace{-4mm}
\section{Experiments}
\vspace{-1mm}
\label{sec:experiments}
\vspace{-2mm}
\input{sections/04_experiments}
\vspace{-3mm}
\section{Discussion}
\vspace{-3mm}
\label{sec:discussion}
\input{sections/05_discussion}

\section*{Acknowledgements}
This worked was supported by the Stanford Human-Centered Artifical Intelligence (HAI) Hoffman-Yee grant, and the NSF Expeditions Grant, Award Number (FAIN) 1918771. We would like to thank the Stanford Center for Research on Foundation Models (CRFM) and Yifan Mei for the tokens and the infrastructure to test different models.

\bibliography{ref}
\bibliographystyle{plainnat}



\appendix

\clearpage
\input{sections/11_appendix}
\end{document}

%% file: sections/00_abstract.tex
As Large Language Models (LLMs) become increasingly integrated into our everyday lives, understanding their ability to comprehend human mental states becomes critical for ensuring effective interactions.
However, despite the recent attempts to assess the Theory-of-Mind (ToM) reasoning capabilities of LLMs, the degree to which these models can align with human ToM remains a nuanced topic of exploration.
This is primarily due to two distinct challenges: (1) the presence of inconsistent results from previous evaluations, and (2) concerns surrounding the validity of existing evaluation methodologies. To address these challenges, we present a novel framework for procedurally generating evaluations \emph{with} LLMs by populating causal templates. Using our framework, we create a new social reasoning benchmark (\textbf{Big}\textbf{ToM}) \emph{for} LLMs which consists of 25 controls and 5,000 model-written evaluations. We find that human participants rate the quality of our benchmark higher than previous crowd-sourced evaluations and comparable to expert-written evaluations. Using \textbf{Big}\textbf{ToM}, we evaluate the social reasoning capabilities of a variety of LLMs and compare model performances with human performance.
Our results suggest that GPT4 has ToM capabilities that mirror human inference patterns, though less reliable, while other LLMs struggle.\footnote{\href{https://sites.google.com/view/social-reasoning-lms}{https://sites.google.com/view/social-reasoning-lms}}


%% file: sections/01_introduction.tex
Humans continually try to understand what others think, want, and feel.

We try to understand what people have done and predict what they might do next by inferring their mental states. This capability, often referred to as ``Theory of Mind'' (ToM), is the foundation of social interaction \citep{wellman1992child, leslie2004core, onishi200515, frith2005theory, spelke2016core}. With Large Language Models (LLMs) playing a growing role in our lives, assessing their ability to model human mental states is key for guaranteeing effective interactions. This involves evaluating the current abilities of LLMs, understanding their failure modes, and discovering ways to improve them.
LLMs with ToM-like abilities could be better at teaching us, learning from us, communicating with us, collaborating with us, and understanding us \citep{hadfield2016cooperative, lake2017building, rabinowitz2018machine, gandhi2021baby, shu2021agent}.

\begin{figure}[!t]
\centering
\includegraphics[width=0.95\textwidth]{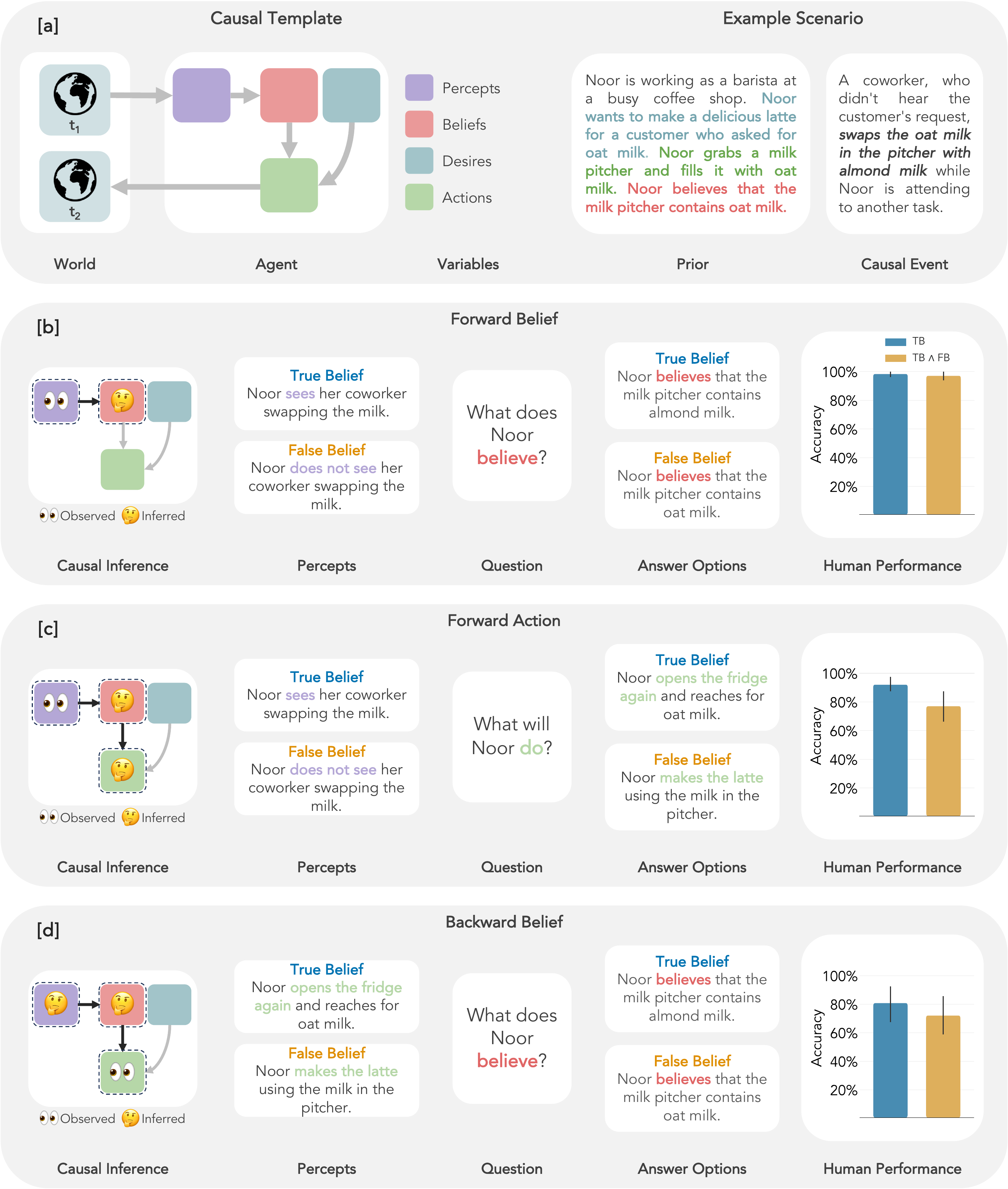}
\caption{Illustration of our template-based Theory-of-Mind (ToM) scenarios. [a] The causal template and an example scenario including prior desires, actions, and beliefs, and a causal event that changes the state of the environment. [b] Testing \textit{Forward Belief} inference by manipulating an agent's percepts. TB = \textcolor{truecolor}{True Belief}. FB = \textcolor{falsecolor}{False Belief}. [c] \textit{Forward Action} inference from an agent's percepts which requires additional inferences over unknown beliefs. [d] \textit{Backward Belief} inference requires joint inferences over unknown percepts and beliefs from an agent's observed actions. Error bars for human performance represent 95\% bootstrapped confidence intervals of the mean.}
\label{fig:fig_1}
\end{figure}

Recent attempts at understanding social reasoning in LLMs have used crowd-sourced data, SocialIQA \citep{sap2019socialiqa}, data from synthetic templates, ToMi \citep{le2019RevisitingTE}, or (modified) tests from psychology designed to evaluate human capabilities \citep[e.g.][]{moghaddam2023boosting, ullman2023large,kosinski2023theory,bubeck2023sparks, ma2023tomchallenges,trott2022large}. 
\citet{sap2022neural} used SocialIQA and ToMi to show that GPT-3 had limitied social reasoning capabilities. However, their findings are challenging to interpret due to limitations in their methodology. SocialIQA has several ambiguous examples and stories that do not effectively test the desired social reasoning behaviors. In comparison, ToMi suffers from ambiguous narratives with unclear perceptual descriptions and additional confounding factors in reasoning like memory loads or tracking requirements.
Moreover, both of these datasets lack control conditions making it difficult to identify precisely where models make mistakes. The results of studies with tests developed by psychologists show some signs of ToM capabilites in LLMs \citep{kosinski2023theory, bubeck2023sparks}. However, when LLMs such as GPT-3 \citep{brown2020language} succeed in scenarios, they often fail dramatically on trivial alterations \citep{ullman2023large,moghaddam2023boosting,shapira2023clever}. Despite their careful design, concerns about the limited test set \citep{moghaddam2023boosting,kosinski2023theory} and potential dataset leakage from modifications to the Sally-Anne task \citep{baron1985does} in \citep{bubeck2023sparks,kosinski2023theory,moghaddam2023boosting}, suggest caution in the interpretation of these results (see \autoref{asec:old-bench} for a detailed discussion).

To address these shortcomings, we present a novel framework for procedurally designing synthetic ToM evaluations from causal templates (\autoref{fig:fig_1}). By representing ToM scenarios as causal graphs, we can systematically intervene on variables, generate control conditions, and probe different aspects of an LLM's ToM capabilities. More concretely, consider the scenario in \autoref{fig:fig_1}a: Here, \textit{``Noor''} is an agent with a \textcolor{desire}{desire}, \textit{``to make a latte with oat milk''}, who performed an \textcolor{action}{action}, \textit{``fills it with oat milk''}, resulting in a \textcolor{belief}{belief}, \textit{``she believes that the pitcher has oat milk''}. Next, a \textit{``Causal Event''} changes the state of the environment (``\textit{oat milk}'' $\rightarrow$ ``\textit{almond milk}''). Given this setup, we can now manipulate the agent's \textcolor{percept}{percept} to create \textcolor{truecolor}{True Belief} and \textcolor{falsecolor}{False Belief} conditions. In the \textcolor{truecolor}{True Belief} condition, the perception of the causal event is presented, \textit{``Noor sees her coworker swapping the milk''}, and then we test a model's \textit{forward belief} inference abilities; \textit{``What does Noor believe is in the pitcher?''} (\autoref{fig:fig_1}b). Moreover, we can probe more difficult inferences, such as \textit{forward action} inferences from an agent's percepts via inferred beliefs (\autoref{fig:fig_1}c). In addition to manipulating percepts, we can intervene on an agent's actions to examine a model's \textit{backward belief} inferences, which is even more difficult as it requires a joint inference over unknown percepts and beliefs (\autoref{fig:fig_1}d; \autoref{sec:methods}).

We design a framework for systematic and diverse evaluations of LLMs in three steps. First, we build a causal template (an abstracted causal model) for the domain of interest, which in our case is ToM. Second, we prompt a language model to populate the variables in the template (yielding a concrete causal model). Third, we construct different evaluation conditions by combining variables from the populated causal template (\autoref{fig:fig_2} and \autoref{sec:methods}). Our approach is a general method for generating evaluations, applicable in any domain where reasoning traces can be represented as causal graphs. 

Overall, our contributions are as follows:
(1) We present a framework for generating systematic evaluations from causal templates that help us understand a model's behavior, its failures and successes, through automated, controlled tests. (2) We show the effectiveness of our scalable, cost-efficient method for writing evaluations with language models by comparing its quality to crowd-sourced and expert written tests. (3) Finally, we test ToM reasoning in a variety of LLMs\footnote{\texttt{LLaMa-65B}, \texttt{text-davinci-003}, \texttt{gpt-3.5-turbo}, \texttt{Claude-v1.3}, \texttt{Claude-2}, \texttt{gpt-4-0314}} using different prompting techniques, and compare model performances with human performance. We find that \texttt{gpt-4} shows human-like ToM inference patterns, although less reliable, while other LLMs struggle.

%% file: sections/02_related-work.tex
\noindent\textbf{Theory-of-Mind in Humans.}
Infants, arguably from 12 months of age, can attribute mental states to agents, exhibiting theory of mind reasoning  \citep{onishi200515}. A classic test to probe this reasoning is the false-belief task \citep{baron1985does}: Sally has a doll and puts it in a basket, then leaves the room.
While Sally is away, Anne takes the ball out of the basket and puts it into a box. Participants are then asked to predict what happens next: ``When Sally comes back, where will she look for her ball?''. To answer this question, participants need to infer Sally's beliefs, and realize that her beliefs aren't the same as theirs. 
Through well-planned experiments, cognitive scientists probe reasoning aspects relating to agents' desires and beliefs \citep{leslie2004core,gergely2003teleological,wellman1992child, spelke2016core}. These studies employ control conditions to rule out simple heuristics people might use, while searching for the cognitive mechanisms that underlie human reasoning and behavior \citep{baker2008theory,baker2017rational,goodman2009cause,wu2023computational,sosa2021dynamics, franken2023naive}. Such experiments have inspired AI researchers to design ``behavioral'' experiments for probing ToM in AI models \citep{gandhi2021baby,shu2021agent,stojnic2023commonsense,kosoy2022learning}.

\noindent\textbf{Theory-of-Mind in Machines.} Initial attempts at building ToM representations in neural network based models \cite{raileanu2018modeling, rabinowitz2018machine} used ToM specific tasks to train and test the models. As LLMs scaled and became better at reasoning, researchers used a small set of tests from cognitive science to claim that ToM reasoning had emerged in LLMs (GPT-3, GPT-4) \citep{kosinski2023theory,bubeck2023sparks}. But, further probing using alterations and diverse scenarios showed that this reasoning was quite brittle \citep{ullman2023large, moghaddam2023boosting}. Other tests for social reasoning used crowd-sourced and synthetic evaluations to find mixed results \citep{shapira2023clever,sap2019socialiqa,sap2019socialiqa,le2019RevisitingTE, trott2022large}. Despite the abundance of research in this domain, we still don't understand the strengths and weaknesses of LLMs in ToM reasoning.
Previous evaluations suffer from one or more of the following issues: reliance on limited evaluations designed for humans \citep[e.g.][]{kosinski2023theory, moghaddam2023boosting}, insufficient control conditions \citep[e.g.][]{sap2022neural, ullman2023large}, limited test cases \citep[e.g.][]{trott2022large, bubeck2023sparks}, noisy/ambiguous crowd-sourced evaluations \citep[e.g.][]{sap2022neural}, the risk of dataset leakage \citep[e.g.][]{kosinski2023theory, trott2022large}, confounding factors in reasoning \citep[e.g.][]{le2019RevisitingTE, ullman2023large}  and possible overfitting of the prompting method \citep{moghaddam2023boosting} (see \autoref{asec:old-bench} for a detailed discussion). The goal of our work is to come up with a scalable, replicable framework to understand the reasoning behind predictions made by language models while avoiding the pitfalls that other methods fall into.

\noindent\textbf{Model-Written Evaluations.}

Advancements in aligning LLMs with instruction-tuning and RL from human feedback (RLHF) have recently shown promising results, such as the generation of a high-quality hate-speech detection dataset with GPT-3 \citep{hartvigsen2022toxigen, efrat2020turking}, red-teaming \citep{perez2022red}, and training data generation \citep{schick2021generating}. The latest work has extended this to the generation of evaluations directly \citep{perez2022discovering}. \citet{perez2022discovering} examined whether generated data can serve as high-quality evaluation data with minimal errors for a variety of novel language model behaviors. These tests, while being scalable, cost-effective and easy to replicate, are still challenging to interpret as they lack structure in the generation of tests. In contrast, \citet{dasgupta2022language} show how carefully designed automated tests can find specific failure modes in reasoning. Our work aims to integrate the benefits of these methods, creating a more structured approach to generating and interpreting tests, while preserving scalability, cost-effectiveness, and ease of replication.

%% file: sections/03_method.tex
\noindent\textbf{Preliminaries.} Theory of Mind is the ability to attribute mental states like beliefs, intents, desires, emotions and knowledge to oneself and others. It involves understanding that other people's mental states (latent causes) guide their actions (see \autoref{fig:fig_1}a). In this work, we focus on the causal graph linking precepts, beliefs, desires, and actions. We want to test if models are able to perform forward and backward inference over different variables in this graph. 

\begin{figure}[t]
\centering
\includegraphics[width=0.99\textwidth]{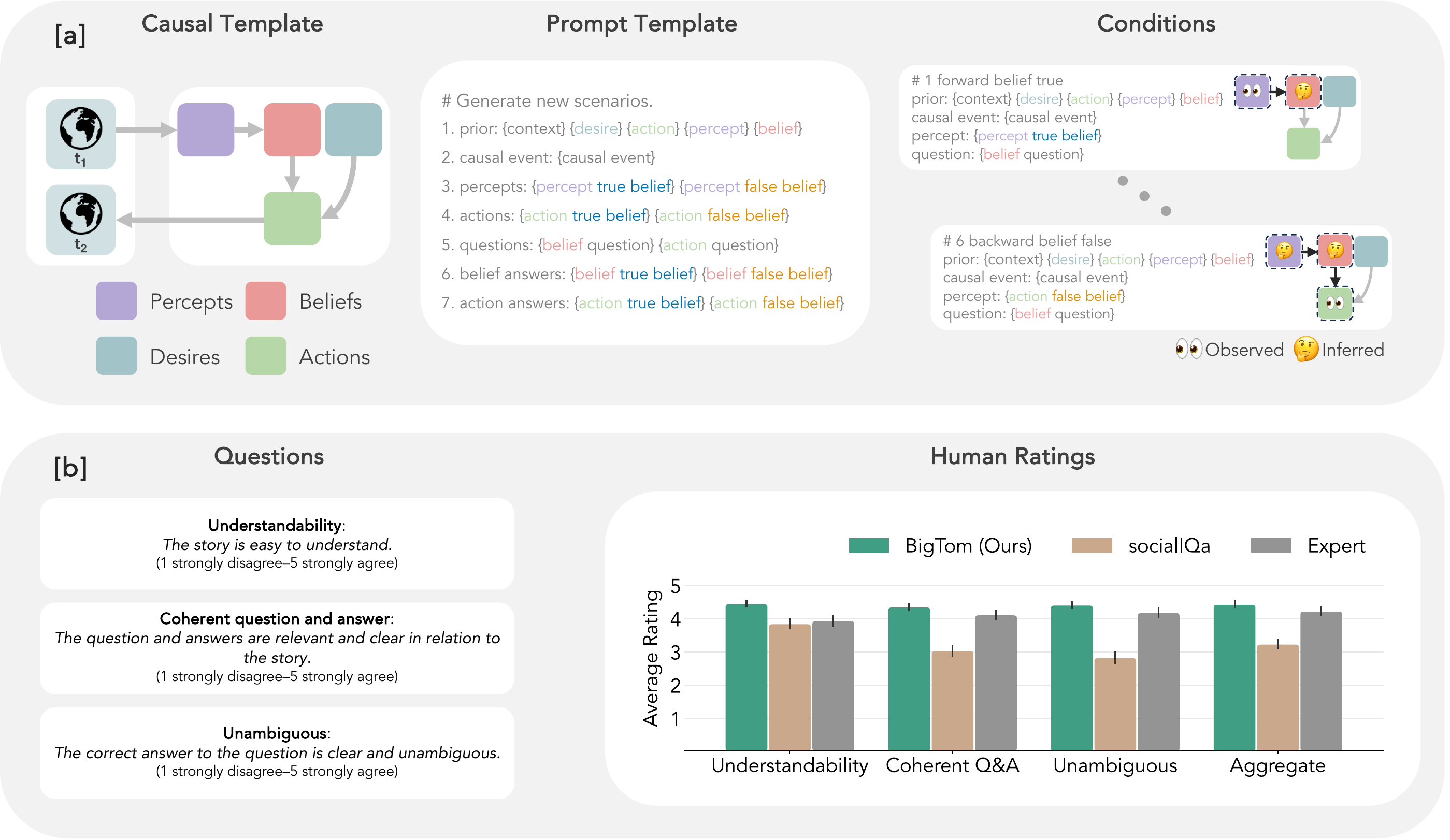}
\caption{[a] Three-stage method for generating evaluations: Building a causal template for the domain (left). Creating a prompt template (simplified here; see \autoref{afig:prompt_for_gen} for the prompt) from the causal graph and populating template variables using a language model (middle). Composing test items by combining template variables (right). [b] Crowdworker ratings of our model-generated Theory-of-Mind (ToM) evaluations compared to crowd-sourced ToM evaluations and expert-written ToM evaluations. Error bars represent 95\% bootstrapped confidence intervals of the mean.}
\label{fig:fig_2}
\end{figure}

Our goal is to generate ToM evaluations that meet the following criteria: (1) they include control conditions to systematically assess language models' response tendencies and failure modes across different aspects of ToM, (2) they don't directly involve human-designed test items, and (3) they are diverse and scalable. By generating a diverse set of tasks, we wish to specifically target the reasoning involved in ToM inferences, while not focusing on other errors in common-sense reasoning\footnote{For example, in \citet{shapira2023clever}, errors in understanding `transparent access' are not ToM inference errors but errors in understanding perceptual access with transparent objects, i.e., not an error in computing what someone knows from what they see. Adding the line: ``<agent> can see through transparent <object>.`` mitigates these errors with \texttt{gpt-4}.}. To achieve this, we follow \citep{perez2022discovering} and propose using language models to generate their own evaluations, specifically story($s$)-question($q$)-answer($a$) test items of the format of $(s_1, q_1, a_1),(s_2, q_2, a_2),...(s_N, q_N, a_N)$ (examples are shown in \autoref{tab:table_1}). To generate these evaluations, we propose a novel three stage-method: (1) Building a causal template of the domain, (2) populating causal templates using language models, and (3) composing test items for a given condition by ``stitching'' together template variables into fluent stories (\autoref{fig:fig_2}a). 

\vspace{-2mm}
\subsection{Stage 1: Building a Causal Template}
\vspace{-2mm}

To build a causal template, we start by defining the variables (see \autoref{fig:fig_1}a and \autoref{fig:fig_2}a). The world is set up with a context and description of the agent (\textit{``Noor is a barista [...]''}). Next, we add the initial (prior) values of the variables in the template: \textcolor{desire}{desire} (\textit{``Noor wants to make a latte``}),  \textcolor{percept}{percept} (\textit{``Noor fills a pitcher with oat milk``}) and \textcolor{belief}{belief} (\textit{``Noor believes that the pitcher has oat milk``}). Next, a \textit{Causal Event} changes the state of the environment (\textit{``oat milk``} $\rightarrow$ \textit{``almond milk''}). We can now manipulate the agent's \textcolor{percept}{percept} of the causal event and the resulting \textcolor{action}{action} the agent will take. In this paper, we focus on the following inferences:

\noindent\textbf{Initial Percept to Initial Belief.} This tests if models understand that percepts (and actions) give rise to beliefs: \textit{``Noor grabs a pitcher and fills it with oat milk``} $\rightarrow$ \textit{``Noor believes that the milk pitcher contains oat milk``}. This is a preliminary inference that a model must perform before being able to answer more complicated questions about beliefs or actions following the causal event. 

\noindent\textbf{With vs. Without Initial Belief.} 
We consider two version of the background (prior) scenario. In version one (\textit{``without initial belief''}), we \underline{do not} explicitly reveal the agent's initial belief (i.e. we exclude the sentence \textit{``Noor believes that the pitcher has oat milk``}). In version two (\textit{``with initial belief''}), we \underline{include} the agent's initial belief in the scenario. Revealing the initial belief should make the inference problem easier as we can skip the inference from percept to belief. Moreover, it allows us to test whether explicitly stating the initial belief biases the answers of LLMs.

\noindent\textbf{Forward Belief.} In this condition, the model must infer the belief of the agent given the agent's percepts of the causal event (see \autoref{fig:fig_1}b). This inference can be written as: $P(\text{Belief} \mid \text{Percept})$. 

\noindent\textbf{Forward Action.} Here, the model must infer the agent's action given percepts (see \autoref{fig:fig_1}c). Implicitly, this inference requires the model to first infer the agent's belief before predicting the agent's action given percept and desire: 
$\sum_{\text{Belief}} P(\text{Action} \mid \text{Percept}, \text{Desire}, \text{Belief})$.


\noindent\textbf{Backward Belief.} In this condition (\autoref{fig:fig_1}d), the goal is to infer the agent's belief from observed actions.
This is the most difficult condition as it requires joint inference over unknown beliefs and percepts from an observed action: 
$\sum_{\text{Percept}} \sum_{\text{Belief}} P(\text{Action} \mid \text{Desire}, \text{Percept}, \text{Belief})$.




\noindent\textbf{Additional Controls.} To control for context effects, we further include a control condition in which the ``Causal Event'' is replaced with a ``Random Event'' that does not change the state of the environment (e.g., \textit{``A musician starts playing music while Noor is making the latte.''}).

\vspace{-2mm}
\subsection{Stage 2: Populating Causal Templates With Language Models}
\vspace{-2mm}
Unlike previous work \citep{perez2022discovering, wang2022self}, we do not directly use language models to generate individual test items. Instead, we create prompt templates (\autoref{fig:fig_2}a, \autoref{asec:method}) from the causal template developed in the previous section and use a language model(\texttt{gpt-4-0314} with a temperature of $0.5$ and default parameters) to fill template variables. For a given prompt, we generate $3$ new completions using $3$ few-shot examples. We constrain the model to generate exactly one sentence for a each variable in our template. Here we make an assumption that the model is good at forward prediction, coming up with plausible actions from the context, and the belief and desire of the agent (see \autoref{asec:failure} for a discussion).

\subsection{Stage 3: Composing Test Items from Template Variables}
\vspace{-2mm}
Having generated a sentence for each variable of the template, we choose the sentences to include in the story; this varies by condition depending on the inferences we wish to test.
For example, we can create a story for the \textit{Forward Belief inference for the True Belief condition} by combining the  sentences for variables \texttt{context}, \texttt{desire}, \texttt{action}, \texttt{percept}, \texttt{belief} with the sentences for \texttt{causal event} and \texttt{percept}, followed by the \texttt{belief question} and the answer options for the \texttt{true belief} and \texttt{false belief} versions (see \autoref{fig:fig_2}a). 
In total, we generate 200 templates and extract 25 conditions from each template (resulting in a new benchmark consisting of 5,000 test items; see \autoref{asec:method} for examples). For our main results with both humans and language models, we will focus on the $6$ most important conditions \textit{Forward Belief} (\textcolor{truecolor}{True Belief}, \textcolor{falsecolor}{False Belief}), \textit{Forward Action} (\textcolor{truecolor}{True Belief}, \textcolor{falsecolor}{False Belief}), and \textit{Backward Belief} (\textcolor{truecolor}{True Belief}, \textcolor{falsecolor}{False Belief}). Results for the remaining conditions are in \autoref{asec:results}.

\begin{table}
  \caption{Test examples from human-written and model-written datasets.}
  \label{tab:table_1}
  \centering
  \small
  \begin{tabular}{lp{10cm}}
    \toprule
    Dataset  &  Human-written or Model-written Test Example\\
    \midrule
    \multirow{2}{*}{\raisebox{-\normalbaselineskip}{\begin{tabular}{@{}c@{}}Expert \end{tabular}}} & \fontsize{7pt}{7pt}\selectfont \textbf{Story}: \textit{``Here is a bag filled with popcorn. There is no chocolate in the bag. Yet, the label on the bag says 'chocolate' and not 'popcorn.' Sam finds the bag. She had never seen the bag before. She cannot see what is inside the bag. She reads the label.''} \textbf{Question}: \textit{``What does Sam believe the bag contains?''} \textbf{Answers}: 
    Popcorn, Chocolate (correct answer). \textbf{Written by}: Human.  \textbf{Source}: \cite{ullman2023large}\normalfont\\

    \multirow{2}{*}{\raisebox{-\normalbaselineskip}{\begin{tabular}{@{}c@{}}Expert \end{tabular}}}  & \fontsize{7pt}{7pt}\selectfont \textbf{Story}: \textit{``Inside the fridge of the Airbnb host's kitchen is a can labeled as 'tuna'. However, the can actually contains sardines and no tuna. Jill, a guest who just arrived at the house, opens the fridge and sees the can. She cannot see what is inside the can, but she reads the label.''} \textbf{Question}: \textit{``What does Jill think is in the can?''} \textbf{Answers}: Sardines, Tuna (correct answer).\textbf{Written by}: Human. \textbf{Source}: \cite{kosinski2023theory} \normalfont\\
    

    \midrule
    
    \multirow{4}{*}{\raisebox{-\normalbaselineskip}{\begin{tabular}{@{}c@{}} BigToM \end{tabular}}} & \fontsize{7pt}{7pt}\selectfont \textbf{Story}: \textit{``Noor is working as a barista at a busy coffee shop. Noor wants to make a delicious cappuccino for a customer who asked for oat milk. Noor grabs a milk pitcher and fills it with oat milk. Noor believes that the milk pitcher contains oatmilk. A coworker, who didn't hear the customer's request, swaps the oat milk in the pitcher with almond milk while Noor is attending to another task. Noor sees her coworker swapping the milk.''} \textbf{Question}: \textit{``Does Noor believe the milk pitcher contains oat milk or almond milk?''} \textbf{Answers}: Noor believes the milk pitcher contains almond milk (correct answer), Noor believes the milk pitcher contains oat milk. \textbf{Written by}: GPT-4. \textbf{Condition}: Forward Belief (True Belief) With Initial Belief.\normalfont\\
    
    \multirow{4}{*}{\raisebox{-\normalbaselineskip}{\begin{tabular}{@{}c@{}} BigToM \end{tabular}}} & \fontsize{7pt}{7pt}\selectfont \textbf{Story}: \textit{``Mei is a pearl diver in a small coastal village in Japan. Mei wants to find a perfect pearl to give to her grandmother for her birthday. Mei spots an oyster at the bottom of the sea that looks to be the right size and age to contain a pearl. Mei believes that the oyster she spotted contains a pearl. A curious octopus opens the oyster, revealing that there is no pearl inside, and then swims away. Mei dives down to collect the oyster.''} \textbf{Question}: \textit{``Does Mei believe the oyster she spotted contains a pearl or that it is empty?''} \textbf{Answers}: Mei believes the oyster she spotted contains a pearl (correct answer), Mei believes the oyster she spotted is empty. \textbf{Written by}: GPT-4. \textbf{Condition}: Backward Belief (False Belief) With Initial Belief.\normalfont\\

    \midrule
 
    \multirow{1}{*}{\raisebox{-\normalbaselineskip}{\begin{tabular}{@{}c@{}}socialIQa \end{tabular}}}  & \fontsize{7pt}{7pt}\selectfont \textbf{Story}: \textit{``Kendall persisted after being told no, and eventually had a positive effect on Lee.''} \textbf{Question}: \textit{``What will Lee want to do next?''} \textbf{Answers}: Refuse to help Kendall, Give into Kendall (correct answer), Give a punch to Kendall's face. \textbf{Written by}: Human. \textbf{Source}: \cite{sap2019socialiqa}\normalfont\\

    \multirow{1}{*}{\raisebox{-\normalbaselineskip}{\begin{tabular}{@{}c@{}}socialIQa \end{tabular}}} & \fontsize{7pt}{7pt}\selectfont \textbf{Story}: \textit{``Lee tried to remain calm when nobody answered the phone call.''} \textbf{Question}: \textit{``What does Lee need to do before this?''} \textbf{Answers}: send a text, try again, pick up the phone (correct answer). \textbf{Written by}: Human. \textbf{Source}: \cite{sap2019socialiqa}\normalfont\\

    \bottomrule
  \end{tabular}
\end{table}

\vspace{-2mm}
\subsection{Quality of Generated Data}
\noindent\textbf{Expert Evaluations.}
\autoref{tab:table_1} shows random examples from human-and model-written datasets. Our model-written examples are high-quality and closely match the pattern of examples generated by human experts. To assess the quality of our model-written dataset, we first had two experts (two authors) independently evaluate $100$ model-written templates including all $25$ conditions ($2500$ test items overall). During their evaluations, experts answered the following questions: \textbf{Question 1}: \textit{``Does the story follow the assigned structure?''} \textbf{Answers}: $1$ (Yes), $0$ (No). \textbf{Question 2}: \textit{``Does the story test the desired behavior?''} \textbf{Answers}: $1$ (\textit{``Strongly Disagree''}) to $5$ (\textit{``Strongly Agree''}). The overall percentage agreement between experts on the first question was 93.94\% with mean ratings of $0.919$ (95\% CI: $0.859$--$0.970$) for expert~1 and $0.960$ (95\% CI: $0.919$--$0.990$) for expert~2. For the second question, average expert ratings were $4.33$ (95\% CI: $4.13$--$4.53$) for expert~1 and 4.35 (95\% CI: $4.18$--$4.52$) for expert~2, both with a median rating of~$5$. 

\noindent\textbf{Participant Evaluations.}
We evaluate the quality of $200$  items from BigToM with human participants\footnote{Preregistration Experiment 1: \href{https://osf.io/qxj2s}{https://osf.io/qxj2s}. Note: We doubled the size of our participants and items upon reviewer's request. All numbers in the preregistration correspond to half of the numbers reported in this paper.}.  Due to the large number of conditions, we gather participant ratings for the true belief and false belief versions of the forward belief condition, as exemplary versions representing the conditions.
 We compare participants' ratings of our model-written evaluations (``\textbf{BigToM}'') with 50 random items sampled from a large-scale (38,000 items), human-written (crowd-sourced) ToM benchmark (``\textbf{socialIQa}'') \citep{sap2019socialiqa} as well as 50 random items sampled from ToM scenarios written by human researchers (``\textbf{Expert}'') \citep[][]{dodell2011fmri, ullman2023large, kosinski2023theory}. Both socialIQa and the Expert test items were selected as they have recently been used to evaluate language models' ToM capabilities \citep[e.g.][]{sap2022neural,ullman2023large, kosinski2023theory, moghaddam2023boosting, shapira2023clever}. \autoref{fig:fig_2}b shows participants' average item ratings for each dataset and question. Our model-written test items (\textbf{BigToM}) received the highest ratings for each question. Results from a Bayesian linear mixed effects regression confirmed that test-items extracted from our model-written templates were better than the crowd-sourced items, particularly in coherence and un-ambiguity, and comparable to (or better than) expert-written test items (details in \autoref{asec:humanone}).

%% file: sections/04_experiments.tex

\noindent\textbf{Evaluating Models.} We test five large language models: \texttt{text-davinci-003},  \texttt{gpt-3.5-turbo},  \texttt{gpt-4-0314}, \texttt{claude-v1.3}, and \texttt{llama-65b-q5} (quantized)\citep{touvron2023llama,Gerganov2023}. All models are used with the most deterministic setting with a temperature of $0$. We test these models with four types of prompts: 0-shot, 0-shot-chain-of-thought \citep{kojima2022large}, 1-shot, and 1-shot-chain-of-thought \citep{wei2022chain}. The example used for the 1-shot prompt is from the \textit{Forward Belief - False Belief} condition, where the inference variable is the belief of the agent. The task is presented to the model in the form of a comprehension question with a story, followed by a question and two answer options. We compare models on their accuracy to answer the questions. We have released our prompts and evaluation scripts on the project page\footnote{\href{https://sites.google.com/view/social-reasoning-lms}{https://sites.google.com/view/social-reasoning-lms}}. We compare models to a human baseline\footnote{Preregistration Experiment 2: \href{https://osf.io/zxw6m}{https://osf.io/zxw6m}} (details in \ref{asec:humantwo}).

\begin{figure}[t]
\centering
\includegraphics[width=0.99\textwidth]{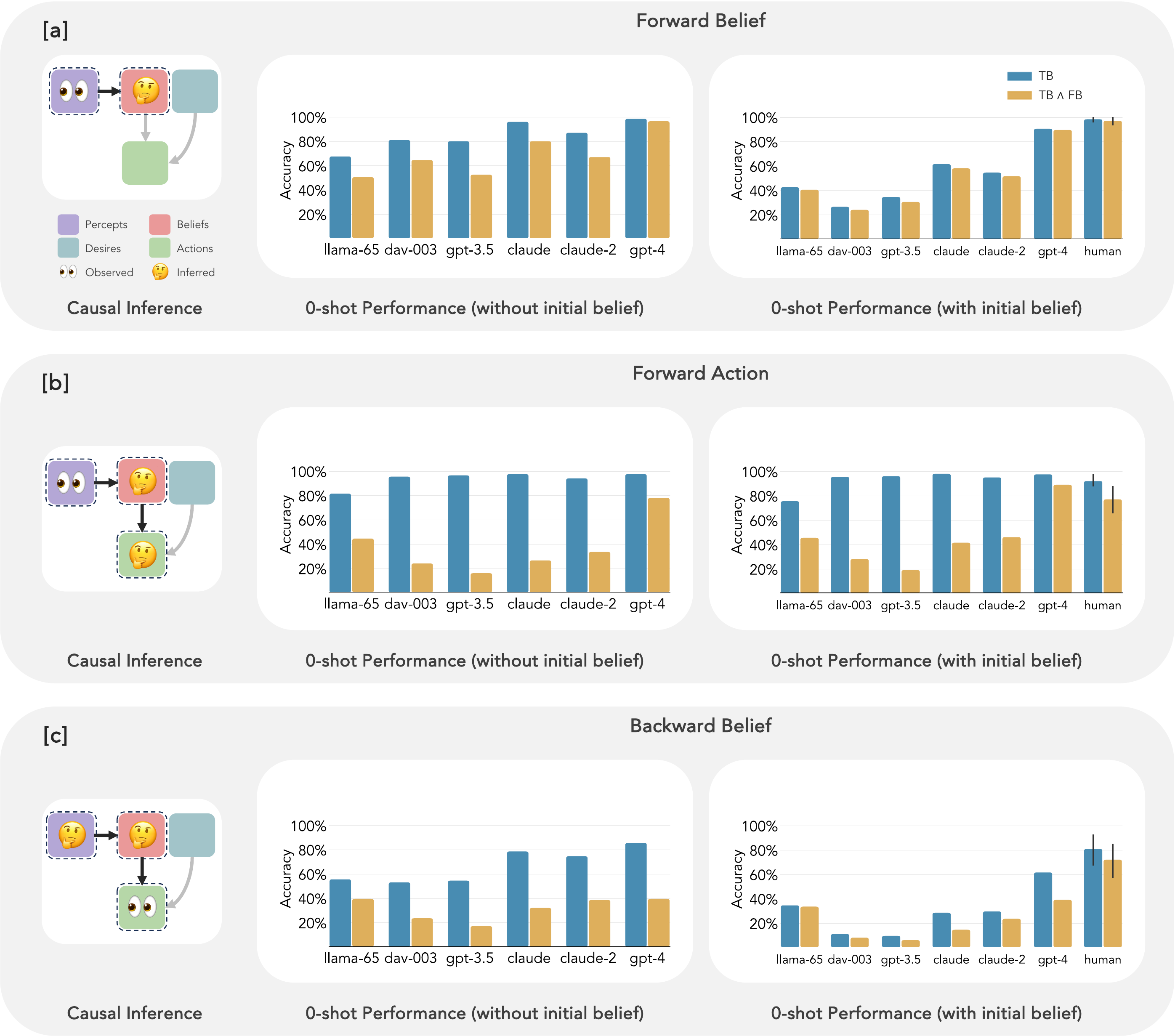}

\caption{{blueModel performance (0-shot) across conditions}. [a] \textit{Forward Belief} inferences from percepts to beliefs. TB = \textcolor{truecolor}{True Belief}. FB = \textcolor{falsecolor}{False Belief}. [b] \textit{Forward Action} inferences from an agent's percepts which require additional inferences over unknown beliefs. [c] \textit{Backward Belief} inferences over unknown percepts and beliefs from an agent's observed actions. Error bars for humans represent 95\% bootstrapped confidence intervals of the mean.}
\label{fig:res}
\end{figure}



\vspace{-2mm}
\subsection{Results and Discussion}
\vspace{-2mm}
The results of our investigation are detailed in \autoref{tab:res}, \autoref{tab:ares} and \autoref{asec:results}, spanning different conditions, models, and prompts.
We discuss results for the true belief and false belief conditions. Importantly, success on the false belief version of the task is evaluated \textit{only} if the model succeeded on the true belief version, as otherwise a model might succeed on the false belief version for the wrong reasons (i.e. failing to comprehend the change in the environment rather than comprehending the change in the environment \textit{and} understanding that the agent was not aware of this change). Therefore, we label the success on the false belief task as ``TB'' $\land$ ``False Belief''. 

\noindent\textbf{Initial Percept to Initial Belief.} All models are proficient at making this inference, and understand how percepts lead to the formation of beliefs (\autoref{asec:results} to table).

\noindent\textbf{Forward Belief Inference.} Here we test if models can track beliefs across the change in the world  (\autoref{tab:res} and \autoref{fig:res}a).
Many models struggle with this, especially when an initial belief is stated (suggesting they anchor on this explicitly stated belief).
\texttt{gpt-4} and, to a lesser extent, \texttt{Claude}  perform better, approaching human levels. 

\noindent\textbf{Forward Action Inference.}While all models are good at predicting actions when beliefs agree with the world state, most models struggle in the critical false-belief condition (\autoref{fig:res}b).
\texttt{gpt-4} is the exception, exhibiting human-level performance (or even slightly better).

\noindent\textbf{Backward Belief Inference.} This represents the most challenging inference. Even humans struggle, achieving only 82\% accuracy in the true belief condition and 72\% in the false belief condition. We believe this is due to unavoidable uncertainty about whether the agent gained knowledge of the true world state.
Models are generally far \emph{below} chance, indicating that they reliably attribute the wrong belief, especially in  false-belief situations and especially when an explicit initial belief is given (\autoref{fig:res}c).
\texttt{gpt-4} is again the exception with a more human-like pattern, though not achieving human level performance 0-shot.


\noindent\textbf{Comparison of prompts.} 
Human participants received instructions and a demonstration example to understand the task (see \autoref{asec:irb}).
Hence, a fair comparison should provide similar support to models.
One-shot learning consistently enhances performance across all models and conditions. In contrast, zero-shot-chain-of-thought (CoT) prompting doesn't consistently improve performance across conditions. Introducing a one-shot CoT example does lead to consistent performance improvement across all conditions, however this performance may not be indicative of stronger ToM \emph{per se}: mimicking the reasoning template is enough to solve our task in most cases. (Human participants were not given demonstrations of how to reason in the task.)


\begin{table}
  \caption{Performance of \textbf{GPT-4} for each method. TB = \textcolor{truecolor}{True Belief}. FB = \textcolor{falsecolor}{False Belief}. $^{\dagger}$ = without initial belief. $^{\ddagger}$ = with initial belief.} 
  \label{tab:res}
  \centering
  \fontsize{8.5pt}{8.5pt}\selectfont
  \begin{tabular}{ccccccc}
    \toprule
    \multicolumn{2}{c}{\multirow{2}{*}{\textbf{Condition}}} 
    & 
    
    \multirow{2}{*}{\textbf{Contingency}} & \multicolumn{4}{c}{\textbf{Method}} \\
  \cline{4-7} 
  \addlinespace[0.5em]
  
  & & & \textit{0-shot} & \textit{0-shot-cot} & \textit{1-shot} & \textit{1-shot-cot} \\
  
\midrule
\addlinespace[0.5em]

\addlinespace[0.5em]
\multirow{3}{*}{\textit{Fwd. Belief}} 
& \raisebox{0.15cm}{\multirow{2}{*}{\includegraphics[width=0.12\textwidth]{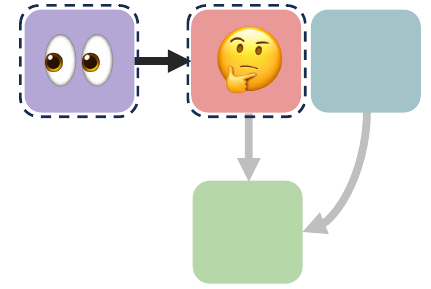}}} & TB & \textcolor{truecolor}{$.99^{\dagger}$ $.91^{\ddagger}$} & \textcolor{truecolor}{$.99^{\dagger}$ $.99^{\ddagger}$} & \textcolor{truecolor}{$.99^{\dagger}$ $.97^{\ddagger}$} & \textcolor{truecolor}{$1.00^{\dagger}$ $.97^{\ddagger}$} \\ 
& & FB & \textcolor{fbonly}{$.98^{\dagger}$ $.99^{\ddagger}$} & \textcolor{fbonly}{$.99^{\dagger}$ $.99^{\ddagger}$} & \textcolor{fbonly}{$.99^{\dagger}$ $.99^{\ddagger}$} & \textcolor{fbonly}{$.99^{\dagger}$ $.99^{\ddagger}$}  \\
&   & TB $\land$ FB & \textcolor{falsecolor}{$.97^{\dagger}$ $.90^{\ddagger}$} & \textcolor{falsecolor}{$.98^{\dagger}$ $.98^{\ddagger}$} & \textcolor{falsecolor}{$.97^{\dagger}$ $.96^{\ddagger}$} & \textcolor{falsecolor}{$.99^{\dagger}$ $.96^{\ddagger}$}  \\
\addlinespace[0.5em]
\cmidrule{1-7}
\addlinespace[0.5em]

\multirow{3}{*}{\textit{Fwd. Action}} 
& \raisebox{0.15cm}{\multirow{2}{*}{\includegraphics[width=0.12\textwidth]{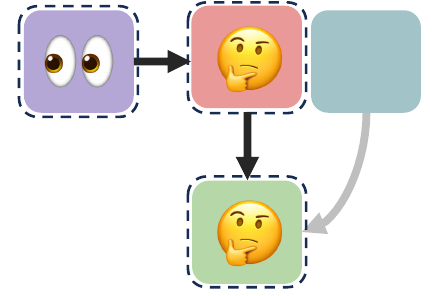}}} & TB& \textcolor{truecolor}{$.98^{\dagger}$ $.98^{\ddagger}$} & \textcolor{truecolor}{$.99^{\dagger}$ $.99^{\ddagger}$} & \textcolor{truecolor}{$1.00^{\dagger}$ $1.00^{\ddagger}$} & \textcolor{truecolor}{$1.00^{\dagger}$ $1.00^{\ddagger}$}  \\ &
& FB & \textcolor{fbonly}{$.81^{\dagger}$ $.92^{\ddagger}$} & \textcolor{fbonly}{$.88^{\dagger}$ $.96^{\ddagger}$} & \textcolor{fbonly}{$.98^{\dagger}$ $1.00^{\ddagger}$} & \textcolor{fbonly}{$1.00^{\dagger}$ $.99^{\ddagger}$}  \\
&                   & TB $\land$ FB & \textcolor{falsecolor}{$.79^{\dagger}$ $.90^{\ddagger}$} & \textcolor{falsecolor}{$.87^{\dagger}$ $.95^{\ddagger}$} & \textcolor{falsecolor}{$.98^{\dagger}$ $1.00^{\ddagger}$} & \textcolor{falsecolor}{$1.00^{\dagger}$ $.99^{\ddagger}$}  \\
\addlinespace[0.5em]
\cmidrule{1-7}
\addlinespace[0.5em]

\multirow{3}{*}{\textit{Bwd. Belief}} 
& \raisebox{0.15cm}{\multirow{2}{*}{\includegraphics[width=0.12\textwidth]{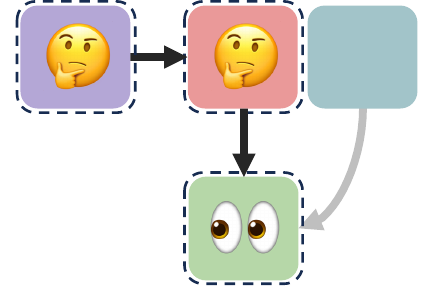}}} & TB& \textcolor{truecolor}{$.86^{\dagger}$ $.62^{\ddagger}$} & \textcolor{truecolor}{$.84^{\dagger}$ $.76^{\ddagger}$} & \textcolor{truecolor}{$.68^{\dagger}$ $.57^{\ddagger}$} & \textcolor{truecolor}{$.83^{\dagger}$ $.81^{\ddagger}$}  \\ &
& FB & \textcolor{fbonly}{$.53^{\dagger}$ $.77^{\ddagger}$} & \textcolor{fbonly}{$.54^{\dagger}$ $.63^{\dagger}$} & \textcolor{fbonly}{$.85^{\dagger}$ $.92^{\ddagger}$} & \textcolor{fbonly}{$.75^{\dagger}$ $.85^{\ddagger}$}  \\
&          & TB $\land$ FB & \textcolor{falsecolor}{$.40^{\dagger}$ $.40^{\ddagger}$} & \textcolor{falsecolor}{$.38^{\dagger}$ $.40^{\ddagger}$} & \textcolor{falsecolor}{$.53^{\dagger}$ $.49^{\ddagger}$} & \textcolor{falsecolor}{$.58^{\dagger}$ $.65^{\ddagger}$}  \\
                 
    \addlinespace[0.5em]
    \bottomrule
  \end{tabular}
\end{table}

%% file: sections/05_discussion.tex
In this work, we present a novel framework for measuring the capabilities of large language models (LLMs) by using abstract causal templates to automatically generate scenarios, controlling what information is provided and what must be inferred. We created a new benchmark for Theory of Mind (BigToM), allowing us to more carefully map the ToM reasoning abilities of LLMs. 
Our extensive control conditions aim to take into account content effects \citep{dasgupta2022language} and many low-level confounds.
We found that many models struggle with even the most basic component of ToM, reasoning from percepts and events to beliefs, and that this is substantially affected by previously-stated beliefs.
Of all the models tested, GPT-4 exhibits ToM capabilities that align closely with human inference patterns.
Yet even GPT-4 was below human accuracy at the most challenging task: inferring beliefs from actions.

\textbf{Limitations.} Our evaluation methodology may appear circular at first: the model being tested plays a role in generating the test items.
However, we believe that for testing inferential abilities this is not a confound.
Our method constructs stories by selecting from all the available facts of a given situation and then isolates the inferential capabilities for the remaining aspects. This means that a model may be able to understand the immediate causal steps in the story while being unable to perform the required inferences being tested.
Indeed, even \texttt{gpt-4} does not achieve a perfect zero-shot score at our tests, indicating this gap between situation knowledge and inferential understanding. Further,  to validate our hypothesis about circularity not being a confound, we generate an evaluation set with \texttt{claude-2}. We find that \texttt{gpt-4} gets comparable scores on an evaluation set generated by a different model, outperforming the model that created the dataset (see \autoref{asec:circ} for details).

Our method shares limitations with other model-generated evaluations (as discussed  in \citet{perez2022discovering}): the generated evaluations can be biased in the content and the contexts that are generated . While synthetic datasets generated from language models offer advantages in scale and control, they also come with inherent biases reflecting those embedded in the underlying model and training data. As large language models are trained on internet text, they inevitably pick up stereotyped associations for gender, race, and other attributes in certain contexts. This could lead to normative, stereotyped roles in different situations in the synthetic dataset. A related issue could arise from biases leading to over-generation of certain situations, effectively yielding imbalanced evaluation data. (We note this is also a problem for human-generated items!) However, language models are also steerable through detailed instructions, allowing mitigation of some biases. Careful steering might be needed during dataset generation to ensure diversity and balance along different dimensions. 
In domains where the models capabilities are lacking, the model will struggle to generate good evaluations. Such limitations could be resolved through shared generation with a human expert while populating the causal graph (see \autoref{asec:method} for an example interface). The stories produced by the model at times exhibit errors in common sense, yet these instances represent a small fraction ($\sim$3\%) of the overall tests generated; as language models continue to improve, we can expect these errors to reduce. Our test items tend to have syntactic similarities which might reduce the diversity of the items in our benchmark; this could perhaps be fixed by asking the model to paraphrase the generated stories.

\textbf{Future Work.} Our causal template method can be used for other domains where the effects of hidden causes or the underlying causes of effects must be inferred. These include many studied by cognitive scientists interested in the ``intuitive theories'' underlying human reasoning. For instance, 
morality, affect, and desire within social cognition, and extending to physical reasoning and more abstract reasoning such as medical diagnosis and mathematics.

In the future, testing social reasoning should move towards more realistic scenarios that are not limited to traditional ToM tests. We believe that we should focus on creating social reasoning tests or benchmarks in use-cases where LLMs are being deployed. We believe that there is a need to move towards more dynamic benchmarks for social reasoning, by creating environments where people or simulated agents (LLMs as people) interact with a language model. Such environments could also be used as a playground where the capabilities of models are not only measured, but also improved.

\textbf{Conclusion.} We have demonstrated a novel approach for assessing LLMs, and while there are limitations, we believe our findings offer a promising direction for future research in understanding and enhancing the capabilities of these powerful models.
The nascent ability of LLMs to reason about mental states of people is a foundational capability for exciting use cases and problematic misuse. Systematic and broad benchmarking of these abilities is thus a pressing concern, and we believe BigToM is an important step.

%% file: sections/11_appendix.tex
\section{Generating Templates}
\label{asec:method}
\textbf{Prompt for generating templates.} See \autoref{afig:prompt_for_gen} for the prompt that we use to generate the template.

\textbf{Example Template.} See \autoref{afig:prompt_template} for an example generation.

\begin{figure}
\centering
\begin{tcolorbox}[
prompt,
title={\small \textbf{Prompt for generating model completions.}},
width=0.9\textwidth]
\fontsize{7pt}{7pt}\selectfont
Generate new stories based on the following template. Be creative and make the stories diverse (from different contexts). Use uncommon names and make the stories different from the examples.\\

Story: \\
1. Context (An agent is in a specific situation or location). \\
2. Prior: Desire (The agent has a specific goal or intention). \\
3. Percept: Perception cues (The agent perceives an object in a specific state without mentioning the agent's knowledge or belief). \\
4. Belief: Belief (The agent believes the object is in the same state mentioned in the previous sentence). \\
5. Causal Event: An external event changes the object's state to another extreme state (without mentioning the agent).\\

Percepts: \\
Aware of event: Agent perceives the event. \\
Not aware of event: Agent does not perceive the event. \\

Actions: \\
Action given new state: An action that the agent anticipates performing in response to the object's altered state. \\
Action given initial state: An action the agent might consider if the object were to persist in its original state. \\

Questions: \\
Belief Question. \\
Desire Question. \\
Action Question.\\

Answers:\\
Belief Answers: \\
Aware: Answer. \\ 
Not Aware: Answer. \\ 
Desire Answers: \\
Aware: Answer. \\ 
Not Aware: Answer. \\ 
Action Answers: \\
Aware: Answer. \\ 
Not Aware: Answer. \\

Random Event.\\

Percepts of random event: \\
Agent is aware of random event. \\
Agent is not aware of random event. 
\normalfont
\end{tcolorbox}
\caption{
Prompt template for generating model completions.
\label{afig:prompt_for_gen}
}
\end{figure}

\begin{figure}
\centering
\begin{tcolorbox}[
prompt,
title={\small \textbf{Example model completion.}},
width=0.9\textwidth]
\fontsize{7pt}{7pt}\selectfont
Story:\\
Context: Mei is a pearl diver in a small coastal village in Japan.\\
Desire: Mei wants to find a perfect pearl to give to her grandmother for her birthday.\\
Perception cues: Mei spots an oyster at the bottom of the sea that looks to be the right size and age to contain a pearl.\\
Belief: Mei believes that the oyster she spotted contains a pearl.\\
An external event: A curious octopus opens the oyster, revealing that there is no pearl inside, and then swims away.\\

Percepts of event:\\
Aware of event: Mei notices the octopus opening the oyster.\\
Not aware of event: Mei does not notice the octopus opening the oyster.\\

Actions:\\
Action given new state: Mei searches for another oyster.\\
Action given initial state: Mei dives down to collect the oyster.\\

Questions:\\
Belief Question: Does Mei believe the oyster she spotted contains a pearl or that it is empty?\\
Desire Question: Does Mei want to collect the oyster she spotted, or does she want to search for another one?\\
Action Question: What will Mei do?\\

Answers given that Agent is Aware of the new object state:\\
Belief Aware: Mei believes the oyster she spotted is empty.\\
Desire Aware: Mei wants to search for another oyster.\\
Action Aware: Mei will search for another oyster.\\

Answers given that Agent is NOT Aware of the new object state:\\
Belief not Aware: Mei believes the oyster she spotted contains a pearl.\\
Desire not Aware: Mei wants to collect the oyster she spotted.\\
Action not Aware: Mei will dive down to collect the oyster.\\

Random Event: A school of brightly colored fish swims past Mei.\\

Percepts of random event:\\
Aware of random event: Mei sees the school of brightly colored fish.\\
Not aware of random event: Mei does not notice the school of brightly colored fish
\normalfont
\end{tcolorbox}
\caption{
Example model completion.
\label{afig:prompt_template}
}
\end{figure}


\begin{figure}[t]
\centering
\includegraphics[width=0.99\textwidth]{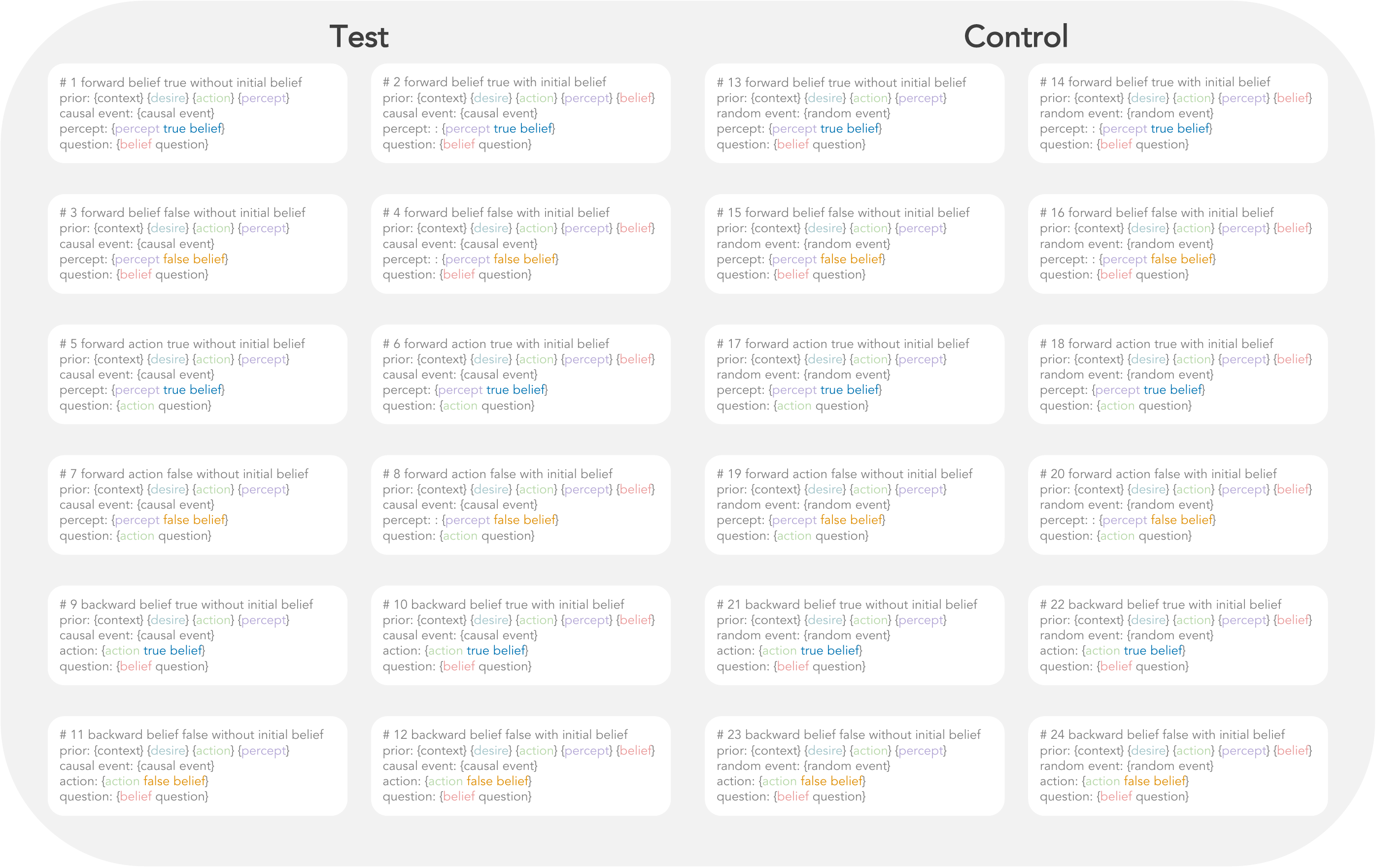}
\caption{All 24 conditions used to evaluate models. Missing from the Figure: Condition 25 which simply included the initial percept followed by a question about the agent's initial belief (to assess whether models can infer beliefs from percepts).}
\label{fig:afig_1}
\end{figure}
\begin{figure}
    \centering
    \includegraphics[width=0.99\textwidth]{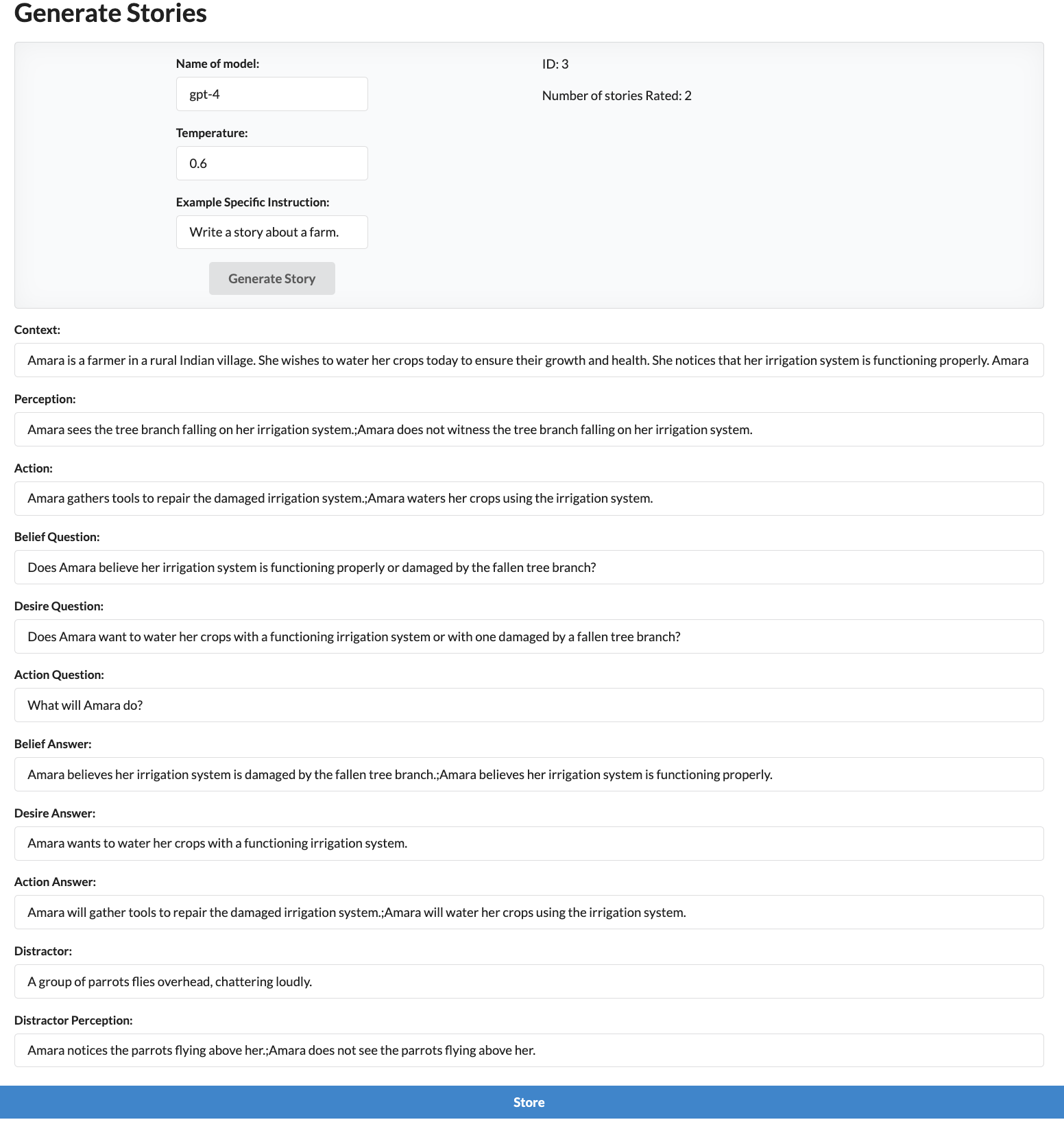}
    \caption{An example interface for shared generation. This interface can be used when the model is not capable enough to populate the template by itself. It should be noted that BigToM is generated fully autonomously. Here, the user can specify the generation parameters and then edit the populated template before storing it.}
    \label{fig:interface}
\end{figure}

\textbf{Interface for Generation.} When a model fails to coherently populate a template for a task, we show how are method can be extended to a semi-autonomous setting. See \autoref{fig:interface} for a sample interface for populating templates with a human-in-the-loop. It must be noted that BigToM is generated fully autonomously. This interface is only shown as an example for domains that are more difficult to generate evaluations for.

\section{Human Experiments}
\label{asec:human}

\subsection{Experiment 1: Human Quality Ratings}
\label{asec:humanone}

\textbf{ Expert dataset.} We choose a size of 50 items as a representative sample of test items written by experts. These were sampled from three sources: 
\begin{itemize}
\item \citet{kosinski2023theory} (21/40 items): This consists of two types of false-belief ToM tasks, widely used in human studies:
    \begin{itemize}
        \item Unexpected Contents Task \citep{perner1987three} (20 items): In this scenario, the protagonist forms a false-belief due to a mislabeled container. The participant must infer that the protagonist incorrectly assumes that the label and its contents are aligned.
        \item Unexpected Transfer Task \citep{wimmer1983beliefs} (20 items): In this scenario, the protagonist witnesses an item in its original state and in the protagonist's absence, the state of this item is changed. The participant must infer that the protagonist still believes that the object is in its original state.
    \end{itemize}
\item \citet{ullman2023large} (9/9 items): The items in this are alterations of the tasks in Kosinski et al.
\item \citet{dodell2011fmri}, \citet{moghaddam2023boosting}: These items are similar to the unexpected transfer task. In addition to stories about false-beliefs, this dataset also contains stories about false/outdated photographs which require the understanding of false/ outdated content.
\end{itemize}

\textbf{Procedure.} We recruited 200 human participants through Prolific \citep{palan2018prolific} and asked each participant to rate 30 items (10 ``Forward Belief True Belief'', 10 ``Forward Belief False Belief'', 5 ``Expert'', 5 ``socialIQa''), resulting in 20 independent participant ratings per item.\footnote{Preregistration Experiment 1: \href{https://osf.io/qxj2s}{https://osf.io/qxj2s}}

\textbf{Results.} \autoref{fig:fig_2}b shows participants' average item ratings for each dataset and question. Our model-written test items (\textbf{BigToM}) received the highest ratings for each question.
To quantitatively compare ratings between datasets, we created an aggregate rating by taking the mean across the three likert questions for a given item and participant. We then fit a Bayesian linear mixed effects model in R included a fixed effect for the datasets and random effects for both items and participants. 
After fitting the model, we computed contrasts between the different datasets and found that for the contrast ``BigToM-socialIQa'' , the estimate was $1.152$ (95\% CI: $1.066$-$1.244$). For ``BigToM-Expert'', the estimate was $0.263$ (95\% CI: $0.178$ -$0.347$), and for ``socialIQa-Expert'', the estimate was $-0.889$ (95\% CI: $-1.000$, $-0.782$). Overall, these results confirmed that test-items extracted from our model-written templates were better than the crowd-sourced items and comparable (or better) than expert-written test items. Contrasts for each dependent measure can be found in \autoref{tab:acontrasts}.

\begin{table} [!htp]
  \caption{Posterior contrasts including 95\% credible intervals for each dependent measure from our human quality evaluations. Aggregate = mean across the three dependent measures.} 
  \label{tab:acontrasts}
  \centering
  \fontsize{8.5pt}{8.5pt}\selectfont
  \begin{tabular}{cccc}
    \toprule
    \multirow{2}{*}{\textbf{Dependent Measure}} & \multicolumn{3}{c}{\textbf{Contrast}} \\
  \cline{2-4} 
  \addlinespace[0.5em]
  & \textit{BigTom-socialIQa} & \textit{BigTom-Expert} & \textit{socialIQ-Expert} \\
  
\midrule
\addlinespace[0.5em]
\multirow{1}{*}{Understandability} 
& $0.503_{[0.424,0.581]}$  & $0.248_{[0.168,0.321]}$ & $-0.254_{[-0.346,-0.1561]}$ \\ 
\addlinespace[0.5em]
\multirow{1}{*}{Coherent Question \& Answer} 
& $1.406_{[1.311,1.496]}$  & $0.203_{[0.117,0.291]}$ & $-1.205_{[-1.315,-1.083]}$ \\ 
\addlinespace[0.5em]
\multirow{1}{*}{Unambiguous}
& $1.541_{[1.412,1.675]}$  & $0.327_{[0.202,0.458]}$ & $-1.214_{[-1.378,-1.057]}$ \\ 
\addlinespace[0.5em]
\multirow{1}{*}{Aggregate} 
& $1.152_{[1.066,1.244]}$ & $ 0.263_{[0.178,0.347]}$ & $-0.889_{    [-1.000,-0.782]}$ \\ 
    \addlinespace[0.5em]
    \bottomrule
  \end{tabular}
\end{table}

\begin{table}
  \caption{Source-wise quality ratings for the expert dataset.} 
  \label{tab:aexpertratings}
  \centering
  \fontsize{8.5pt}{8.5pt}\selectfont
  \begin{tabular}{cccc}
    \toprule
    \multirow{1}{*}{\textbf{Data Source}} & \textit{Aggregate} & \textit{Standard Deviation} \\
  
\midrule
\addlinespace[0.5em]
\multirow{1}{*}{Dodell-Feder} \citep{dodell2011fmri}
& $4.087$  & $0.907$ \\ 
\addlinespace[0.5em]
\multirow{1}{*}{Kosinski} \citep{kosinski2023theory}
& $4.419$  & $0.763$ \\ 
\addlinespace[0.5em]
\multirow{1}{*}{Ullman} \citep{ullman2023large}
& $3.812$  & $0.867$ \\ 
\addlinespace[0.5em]
    \bottomrule
  \end{tabular}
\end{table}

\subsection{Experiment 2: Human Performance}
\label{asec:humantwo}
For the human baseline for the three main conditions (version: \textit{``with initial belief''}), we recruited 20 participants through Prolific.\footnote{Preregistration Experiment 2: \href{https://osf.io/zxw6m}{https://osf.io/zxw6m}} Participants were paid \$$12.05$/hr. 

\section{Failure cases}
\label{asec:failure}
We tried generating our templates using \texttt{text-davinci-003} and \texttt{gpt-3.5-turbo}. We found that these models were worse at following instructions and frequently made common sense errors. To mitigate this problem we tried to split the generation process into more stages: generating a context, generating initial states, beliefs and desires, generating a causal event and then stitching the story together. Although this approach yielded improvement, the reliability of the generated content still fell short of our expectations. To further improve generations, we added verifiers (the model judging \& giving feedback on the populated template) and revisers (the model revising a template based on feedback) to the pipeline. This improved the quality of generations but the number of mistakes made by the model were still high. Switching to \texttt{gpt-4} with a single generation stage gave high quality generations quite reliably. \texttt{gpt-4} occasionally fails to follow the structure of the template about 2-3\% of the times. Verifying this structure is easy while parsing the template. We simply reject the templates that don't follow the assigned structure. \texttt{gpt-4} made  common-sense errors (1-2\%) in cases where the agent's \textit{awareness of the change in state} in the environment was incorrect.

\section{Previous Benchmarks for ToM Reasoning in LLMs}
\label{asec:old-bench}
\begin{table}
  \caption{Test examples from different social reasoning benchmarks.}
  \label{atab:datasets}
  \centering
  \small
  \begin{tabular}{lp{10cm}}
    \toprule
    Dataset  &  Human-written or Model-written Test Example\\
    \midrule
    \multirow{2}{*}{\raisebox{-\normalbaselineskip}{\begin{tabular}{@{}c@{}}ADV-CFMB \citep{ullman2023large, shapira2023clever} \end{tabular}}} & \fontsize{7pt}{7pt}\selectfont \textbf{Story}: \textit{``On the shelf, there is a transparent bottle. It is full of beer; there is no orange juice in it. Yet, the label on this bottle says "orange juice" and not "beer". Mark walks into the room and notices the bottle. He has never seen it before. He reads the label.''} \textbf{Question}: \textit{``What does he believe the bottle is full of?''} \textbf{Answers}: 
    Beer (correct answer), Orange Juice . \textbf{Source}: \citet{shapira2023clever}\normalfont\\

    \multirow{2}{*}{\raisebox{-\normalbaselineskip}{\begin{tabular}{@{}c@{}}\citet{kosinski2023theory} \end{tabular}}}  & \fontsize{7pt}{7pt}\selectfont \textbf{Story}: \textit{``Inside the fridge of the Airbnb host's kitchen is a can labeled as 'tuna'. However, the can actually contains sardines and no tuna. Jill, a guest who just arrived at the house, opens the fridge and sees the can. She cannot see what is inside the can, but she reads the label.''} \textbf{Question}: \textit{``What does Jill think is in the can?''} \textbf{Answers}: Sardines, Tuna (correct answer). \textbf{Source}: \citet{kosinski2023theory} \normalfont\\
    
    \multirow{2}{*}{\raisebox{-\normalbaselineskip}{\begin{tabular}{@{}c@{}} \citet{ moghaddam2023boosting}\end{tabular}}}  & \fontsize{7pt}{7pt}\selectfont \textbf{Story}: \textit{``The morning of the high school dance Sarah placed her high heel shoes under her dress and then went shopping. That afternoon, her sister borrowed the shoes and later put them under Sarah's bed.''} \textbf{Question}: \textit{``When Sarah gets ready, does she assume her shoes are under her dress?''} \textbf{Answers}: Yes (correct answer), No. \textbf{Source}: \citet{dodell2011fmri}\normalfont   \\

    \multirow{2}{*}{\raisebox{-\normalbaselineskip}{\begin{tabular}{@{}c@{}} ToMi \citep{le2019RevisitingTE}\end{tabular}}}  & \fontsize{7pt}{7pt}\selectfont \textbf{Story}:\textit{``1 Oliver dislikes the kitchen 2 Carter entered the porch. 3 Abigail entered the porch. 4 The potato is in the green suitcase. 5 Abigail exited the porch. 6 Abigail entered the hall. 7 Carter moved the potato to the green envelope. 8 Oliver entered the hall''} \textbf{Question:} \textit{``Where will Abigail look for the potato?''}\textbf{Answers}: green suitcase (correct answer), green envelope. \textbf{Source}: \citet{le2019RevisitingTE}    \normalfont\\

    \multirow{1}{*}{\raisebox{-\normalbaselineskip}{\begin{tabular}{@{}c@{}}socialIQa \citep{sap2019socialiqa} \end{tabular}}}  & \fontsize{7pt}{7pt}\selectfont \textbf{Story}: \textit{``Kendall persisted after being told no, and eventually had a positive effect on Lee.''} \textbf{Question}: \textit{``What will Lee want to do next?''} \textbf{Answers}: Refuse to help Kendall, Give into Kendall (correct answer), Give a punch to Kendall's face. \textbf{Source}: \citet{sap2019socialiqa}\normalfont\\

    \multirow{1}{*}{\raisebox{-\normalbaselineskip}{\begin{tabular}{@{}c@{}}socialIQa \citep{sap2019socialiqa}\end{tabular}}} & \fontsize{7pt}{7pt}\selectfont \textbf{Story}: \textit{``Lee tried to remain calm when nobody answered the phone call.''} \textbf{Question}: \textit{``What does Lee need to do before this?''} \textbf{Answers}: send a text, try again, pick up the phone (correct answer).  \textbf{Source}: \citet{sap2019socialiqa}\normalfont\\

    \bottomrule
  \end{tabular}
\end{table}
See \autoref{atab:datasets} for examples from different datasets.

\textbf{ToMi.} 
ToMi \citep{le2019RevisitingTE}, utilizes templates to generate theory of mind queries akin to those seen in the Sally-Anne tasks. However, the scope of these tasks is fairly narrow, being limited to modifications in object locations.  The perceptual access of different agents in the scene is not clearly defined. In several cases, the stories in ToMi are also ambiguous. Additionally, ToMi has several factors that potentially interfere with the accurate assessment of the theory of mind. These factors involve demands on memory and tracking; the questions posed are extensive, necessitating the simultaneous tracking of multiple locations and agents.  Finally, a lack of control conditions makes some evalutations with this dataset difficult to interpret.

\textbf{SocialIQA.} These tests have been crowd-sourced and tend to be quite noisy with very ambiguous answers. Several questions in the dataset don't test the desired behaviors. The lack of any structure to the dataset makes evaluations with this benchmark difficult to interpret.

\textbf{\citet{bubeck2023sparks, kosinski2023theory}} Tests in these works have a high similarity with other false-belief tasks that have been used with humans. Success on these could merely be the result of dataset leakage. Trivial alterations of the tasks, as shown by \citet{ullman2023large} lead to failure. Finally, the number of test examples in these benchmarks is quite low, making it difficult to systematically test capabilities.

\textbf{\citet{moghaddam2023boosting}} Tests in \citet{moghaddam2023boosting} (from \citet{dodell2011fmri}) face limitations similar to \citet{bubeck2023sparks, kosinski2023theory}. \citet{moghaddam2023boosting} show that one-shot CoT leads to a perfect score on their set of tests. These results are difficult to interpret as all questions require a similar reasoning structure to solve them. So, success with the one-shot CoT example may reduce the generalization gap to an extent that theory of mind capabilities aren't actually being tested.

\textbf{\citet{ullman2023large,shapira2023clever}} \citet{ullman2023large} introduce alterations of tests from \citet{kosinski2023theory} to show how models make mistakes with theory of mind inferences. Based on this dataset, \citet{shapira2023clever} introduce a new benchmark, ADVersarial CommonSense with False-Belief (Adv-CSFB) with a larger number of test examples to test for theory-of-mind. Models, including \texttt{gpt-4}, fair poorly on this benchmark. We believe that failures on this benchmark are due to a lack of information about perceptual access, i.e., failures in understanding how some situations change perceptual access, and not failure in inferences related to theory of mind. For example, adding the line ``<agent> can see through transparent <object>'' leads to success on tasks with transparent access with \texttt{gpt-4}. With problems relating to uninformative labels, where the label is in a different language, adding the line ``<agent> cannot read <differnt language>.'' leads to success on the task.


\begin{table}
  \caption{Five additional example test items from our model-written dataset.}
  \label{tab:aexamples}
  \centering
  \small
  \begin{tabular}{lp{10cm}}
    \toprule
    Dataset  &  Model-written Test Example\\
    \midrule
    
    \multirow{4}{*}{\raisebox{-\normalbaselineskip}{\begin{tabular}{@{}c@{}} BigToM \end{tabular}}} & \fontsize{7pt}{7pt}\selectfont \textbf{Story}: \textit{``Rashid is a hiker who has set out to climb a challenging mountain trail. His goal is to reach the summit before sunset to enjoy the panoramic view. Rashid observes that the weather is clear and sunny, perfect for hiking. Rashid believes that the weather conditions are ideal for his hike. As Rashid continues his ascent, a dense fog suddenly rolls in, reducing visibility on the trail. Rashid does not notice the dense fog.''} \textbf{Question}: \textit{``What will Rashid do?''} \textbf{Answers}: Rashid will confidently hike towards the summit (correct answer), Rashid will consider waiting for the fog to clear before continuing his hike. \textbf{Written by}: GPT-4. \textbf{Condition}: Forward Action (False Belief) With Initial Belief.\normalfont\\
    
    \multirow{4}{*}{\raisebox{-\normalbaselineskip}{\begin{tabular}{@{}c@{}} BigToM \end{tabular}}} & \fontsize{7pt}{7pt}\selectfont \textbf{Story}: \textit{``Aarav is a landscape architect working on a new public park. He wants to plant a row of trees that will provide shade and beauty to the park's visitors. Aarav sees that the saplings he has ordered appear to be strong and healthy. However, an unexpected heatwave hits the area, causing the saplings to wilt and weaken. Aarav does not notice the heatwave's effects on the saplings.''} \textbf{Question}: \textit{``Does Aarav believe the saplings are strong and healthy or wilted and weakened?''} \textbf{Answers}: Aarav believes the saplings are strong and healthy (correct answer), Aarav believes the saplings are wilted and weakened. \textbf{Written by}: GPT-4. \textbf{Condition}: Forward Belief (False Belief) Without Initial Belief.\normalfont\\

     \multirow{4}{*}{\raisebox{-\normalbaselineskip}{\begin{tabular}{@{}c@{}} BigToM \end{tabular}}} & \fontsize{7pt}{7pt}\selectfont \textbf{Story}: \textit{``Naveen is a professional baker who runs a small bakery in a bustling city. Today, he plans to bake a large batch of chocolate chip cookies for an upcoming event. Naveen notices that the oven is preheated to the correct temperature. He believes that the oven is ready for baking. Naveen listens to the street musician's performance. Naveen starts baking the chocolate chip cookies.''} \textbf{Question}: \textit{``Does Naveen believe the oven is preheated to the correct temperature or affected by the power outage and cooled down?''} \textbf{Answers}: Naveen believes the oven is preheated to the correct temperature (correct answer), Naveen believes the oven is affected by the power outage and has cooled down. \textbf{Written by}: GPT-4. \textbf{Condition}: Backward Belief Control (False Belief) With Initial Belief.\normalfont\\

      \multirow{4}{*}{\raisebox{-\normalbaselineskip}{\begin{tabular}{@{}c@{}} BigToM \end{tabular}}} & \fontsize{7pt}{7pt}\selectfont \textbf{Story}: \textit{``Dalia is a professional photographer, setting up her equipment for a wedding photoshoot. She wants to capture stunning photos with her high-quality camera. Dalia sees that the camera lens appears clean and free of smudges. Dalia believes that the camera lens is clean and free of smudges. While Dalia is away, a mischievous child plays with the camera, leaving fingerprints all over the lens. Dalia notices the fingerprints on the camera lens.''} \textbf{Question}: \textit{``What will Dalia do?''} \textbf{Answers}: Dalia will carefully clean the lens before starting the photoshoot (correct answer), Dalia will begin the photoshoot with her clean camera lens. \textbf{Written by}: GPT-4. \textbf{Condition}: Forward Action (True Belief) With Initial Belief.\normalfont\\

       \multirow{4}{*}{\raisebox{-\normalbaselineskip}{\begin{tabular}{@{}c@{}} BigToM \end{tabular}}} & \fontsize{7pt}{7pt}\selectfont \textbf{Story}: \textit{``Azola is a teacher at a small school in rural China. She wants to surprise her students with a new book for the school library. Azola got a notification from Amazon stating that the package containing the book was left at her doorstep. Azola believes the book she ordered has arrived in the package at her doorstep. A gust of wind blows the package off her doorstep, and a neighbor replaces it with a different package containing a hand-knit scarf. Azola retrieves the original package with the book.''} \textbf{Question}: \textit{``Does Azola believe the package contains the book she ordered or a hand-knit scarf?''} \textbf{Answers}: Azola believes the package contains a hand-knit scarf (correct answer), Azola believes the package contains the book she ordered. \textbf{Written by}: GPT-4. \textbf{Condition}: Backward Belief (True Belief) With Initial Belief.\normalfont\\

    \bottomrule
  \end{tabular}
\end{table}


\clearpage
\section{Evaluating models}
\label{asec:results}
\textbf{Resources.} The resources used for our experiments include the APIs associated with the models; in particular the OpenAI API and the Anthropic API. The llama model was evaluated on an internal cluster using a single NVIDIA A40 GPU.

\textbf{Prompts.} See \autoref{fig:eval_prompt0} for the 0-shot prompt, \autoref{fig:eval_prompt0cot} for the 0-shot CoT prompt,\autoref{fig:eval_prompt1} for the 1-shot prompt,\autoref{fig:eval_prompt1cot} for the 1-shot CoT prompt.

\begin{figure}
\centering
\begin{tcolorbox}[
prompt,
title={\small \textbf{Instructions Experiment 1.}},
width=0.9\textwidth]
\fontsize{7pt}{7pt}\selectfont
Please read these instructions carefully, as understanding them is important for successfully completing the study. After reading the instructions, you will need to complete a comprehension quiz to confirm that you understood the task.\\

Instructions:\\

In this online experiment, your task is to evaluate the quality of different short social reasoning scenarios. Each scenario consists of a story, a question, and different answer options. We are interested in whether the story is easy to understand and whether the question and answers make sense with respect to the story. We also want to know whether the 'correct' answer to the question is unclear or ambiguous.\\

An example scenario can look like this:\\

\#\#\#\#\\

\{example 1\}\\

\#\#\#\#\\

Another example of a short story can look like this:\\

\#\#\#\#\\

\{example 2\}\\

\#\#\#\#\\

In total, you will evaluate 30 scenarios. It is important to read each of them carefully and respond to the best of your ability.\\
\normalfont
\end{tcolorbox}
\caption{
Instructions shown to participants in Experiment 1.
\label{fig:afig_instructions_exp_1a}
}
\end{figure}

\begin{figure}
\centering
\begin{tcolorbox}[
prompt,
title={\small \textbf{Evaluation Prompt 0-shot.}},
width=0.9\textwidth]
\fontsize{7pt}{7pt}\selectfont
Answer the questions based on the context. Keep your answer concise, few words are enough, maximum one sentence. Answer as 'Answer:<option>)<answer>'
\normalfont
\end{tcolorbox}
\caption{
Prompts used to evaluate the model. For chat based models, the instructions are presented as a system message and the questions are presented as a user message.
\label{fig:eval_prompt0}
}
\end{figure}

\begin{figure}
\centering
\begin{tcolorbox}[
prompt,
title={\small \textbf{Evaluation Prompt 0-shot CoT.}},
width=0.9\textwidth]
\fontsize{7pt}{7pt}\selectfont
\# Text Models
\\\\
Answer the questions based on the context. Think through your reasoning before answering in 'Thought:'. Write your answer as Answer:<option>)<answer>'.
\\\\
\# Chat Models
\\\\
Answer the questions based on the context. Reason step by step before answering in 'Thought: Let's think step by step'. Write your final answer as 'Answer:<option>)<answer>'. Always pick an option, do not say none of the above or that there is not enough information.
\normalfont
\end{tcolorbox}
\caption{
Prompts used to evaluate the model. For chat based models, the instructions are presented as a system message and the questions are presented as a user message.
\label{fig:eval_prompt0cot}
}
\end{figure}

\begin{figure}
\centering
\begin{tcolorbox}[
prompt,
title={\small \textbf{Evaluation Prompt 1-shot.}},
width=0.9\textwidth]
\fontsize{7pt}{7pt}\selectfont
Answer the questions based on the context. Keep your answer concise, few words are enough, maximum one sentence. Answer as 'Answer:<option>)<answer>'

Story: Kofi is a fisherman from a small village in Ghana. He wants to catch enough fish today to provide for his family and sell the surplus at the market. Kofi repaired his fishing net last night. While Kofi is away from his boat, a group of monkeys comes and plays with the fishing net, tearing it apart. Kofi does not see the monkeys damaging his fishing net

Question: Does Kofi believe his fishing net is in good condition or torn apart?

Choose one of the following:

a)Kofi believes his fishing net is in good condition.

b)Kofi believes his fishing net is torn apart.
\normalfont
\end{tcolorbox}
\caption{
Prompts used to evaluate the model. For chat based models, the instructions are presented as a system message and the questions are presented as a user message.
\label{fig:eval_prompt1}
}
\end{figure}

\begin{figure}
\centering
\begin{tcolorbox}[
prompt,
title={\small \textbf{Evaluation Prompt 1-shot CoT.}},
width=0.9\textwidth]
\fontsize{7pt}{7pt}\selectfont
Answer the questions based on the context. Think through your reasoning before answering in 'Thought:'. Write your answer as Answer:<option>)<answer>'.

Story: Kofi is a fisherman from a small village in Ghana. He wants to catch enough fish today to provide for his family and sell the surplus at the market. Kofi repaired his fishing net last night. While Kofi is away from his boat, a group of monkeys comes and plays with the fishing net, tearing it apart. Kofi does not see the monkeys damaging his fishing net

Question: Does Kofi believe his fishing net is in good condition or torn apart?

Choose one of the following:

a)Kofi believes his fishing net is in good condition.

b)Kofi believes his fishing net is torn apart.

Thought: Let\'s think step by step:

1) Kofi repaired his fishing net last night. So last night he believes that his net is fixed.

2) While Kofi is away from his boat, a group of monkeys comes and plays with the fishing net, tearing it apart.

3) Kofi does not see the monkeys damaging his fishing net. So, his belief about his net stays the same. He thinks that it is fixed.

4) Does Kofi believe his fishing net is in good condition or torn apart?

5) Kofi believes his fishing net is in good condition.

Answer: a)Kofi believes his fishing net is in good condition.
\normalfont
\end{tcolorbox}
\caption{
Prompts used to evaluate the model. For chat based models, the instructions are presented as a system message and the questions are presented as a user message.
\label{fig:eval_prompt1cot}
}
\end{figure}

\textbf{Results of models.} See \autoref{tab:ares} for the results of all models across different conditions.

\textbf{Results on Controls.} See \autoref{tab:arescontrol} for results of the model on the control conditions.

\begin{table}
  \caption{{Model performance for each method}. TB = \textcolor{truecolor}{True Belief}. FB = \textcolor{falsecolor}{False Belief}. $^{\dagger}$ = without initial belief. $^{\ddagger}$ = with initial belief.}
  \label{tab:ares}
  \centering
  \fontsize{6.5pt}{6.5pt}\selectfont
  \begin{tabular}{lcccccc}
    \toprule
    \multirow{2}{*}{\textbf{Model}} & \multirow{2}{*}{\textbf{Condition}} & \multirow{2}{*}{\textbf{Contingency}} & \multicolumn{4}{c}{\textbf{Method}} \\
  \cline{4-7} 
  \addlinespace[0.5em]
  
  & & & \textit{0-shot} & \textit{0-shot-cot} & \textit{1-shot} & \textit{1-shot-cot} \\
  
\midrule
\addlinespace[0.5em]
\addlinespace[0.5em]
\multirow{11}{*}{\textbf{llama-65}} & 
\raisebox{0.15cm}{\multirow{2}{*}{\includegraphics[width=0.065\textwidth]{figures/table_fwd_belief.pdf}}} & TB & 
\textcolor{truecolor}{$.68^{\dagger}$  $.43^{\ddagger}$} & \textcolor{truecolor}{$.40^{\dagger}$ $.34^{\ddagger}$} & \textcolor{truecolor}{$.82^{\dagger}$ $.62^{\ddagger}$} & \textcolor{truecolor}{$.42^{\dagger}$ $.31^{\dagger}$} \\ &
& FB & \textcolor{fbonly}{$.62^{\dagger}$ $.72^{\ddagger}$} & \textcolor{fbonly}{$.77^{\dagger}$ $.74^{\ddagger}$} & \textcolor{fbonly}{$.82^{\dagger}$ $.89^{\ddagger}$} & \textcolor{fbonly}{$.75^{\dagger}$ $.75^{\dagger}$}  \\
&          \fontsize{8pt}{8pt}\selectfont \textit{Fwd. Belief} \normalfont               & TB $\land$ FB & \textcolor{falsecolor}{$.51^{\dagger}$ $.41^{\ddagger}$} & \textcolor{falsecolor}{$.34^{\dagger}$ $.31^{\ddagger}$} & \textcolor{falsecolor}{$.71^{\dagger}$ $.60^{\ddagger}$} & \textcolor{falsecolor}{$.36^{\dagger}$ $.28^{\dagger}$}  \\

\cmidrule{2-7}
\addlinespace[0.5em]

& \raisebox{0.15cm}{\multirow{2}{*}{\includegraphics[width=0.065\textwidth]{figures/table_fwd_action.pdf}}} & TB & \textcolor{truecolor}{$.82^{\dagger}$ $.76^{\ddagger}$} & \textcolor{truecolor}{$.31^{\dagger}$ $.31^{\dagger}$} & \textcolor{truecolor}{$.95^{\dagger}$ $.93^{\ddagger}$} & \textcolor{truecolor}{$.34^{\dagger}$ $.27^{\dagger}$}  \\ &
& FB & \textcolor{fbonly}{$.47^{\dagger}$ $.52^{\ddagger}$} & \textcolor{fbonly}{$.19^{\dagger}$ $.43^{\dagger}$} & \textcolor{fbonly}{$.50^{\dagger}$ $.58^{\ddagger}$} & \textcolor{fbonly}{$.20^{\dagger}$ $.39^{\dagger}$} \\
&                \fontsize{8pt}{8pt}\selectfont \textit{Fwd. Action} \normalfont            & TB $\land$ FB & \textcolor{falsecolor}{$.45^{\dagger}$ $.46^{\ddagger}$} & \textcolor{falsecolor}{$.10^{\dagger}$ $.20^{\dagger}$} & \textcolor{falsecolor}{$.49^{\dagger}$ $.55^{\ddagger}$} & \textcolor{falsecolor}{$.10^{\dagger}$ $.15^{\dagger}$}  \\

\cmidrule{2-7}
\addlinespace[0.5em]

& \raisebox{0.15cm}{\multirow{2}{*}{\includegraphics[width=0.065\textwidth]{figures/table_bwd_belief.pdf}}} & TB & \textcolor{truecolor}{$.56^{\dagger}$ $.35^{\ddagger}$} & \textcolor{truecolor}{$.25^{\dagger}$ $.18 ^{\ddagger}$ } & \textcolor{truecolor}{$.53^{\dagger}$ $.31^{\ddagger}$} & \textcolor{truecolor}{$.25^{\dagger}$ $.19^{\dagger}$}  \\ &
& FB & \textcolor{fbonly}{$.53^{\dagger}$ $.73^{\ddagger}$} & \textcolor{fbonly}{$.69^{\dagger}$ $.75^{\ddagger}$} & \textcolor{fbonly}{$.69^{\dagger}$ $.84^{\ddagger}$} & \textcolor{fbonly}{$.66^{\dagger}$ $.71^{\dagger}$}  \\
&        \fontsize{8pt}{8pt}\selectfont \textit{Bwd. Belief} \normalfont                   & TB $\land$ FB & \textcolor{falsecolor}{$.40^{\dagger}$ $.34^{\ddagger}$} & \textcolor{falsecolor}{$.21^{\dagger}$ $.18^{\ddagger}$} & \textcolor{falsecolor}{$.38^{\dagger}$ $.26^{\ddagger}$} & \textcolor{falsecolor}{$.21^{\dagger}$ $.19^{\dagger}$}  \\

\addlinespace[0.5em]
\midrule
\addlinespace[0.5em]

\addlinespace[0.5em]
\multirow{11}{*}{\textbf{dav-003}} 

& \raisebox{0.15cm}{\multirow{2}{*}{\includegraphics[width=0.065\textwidth]{figures/table_fwd_belief.pdf}}} & TB & \textcolor{truecolor}{$.82^{\dagger}$ $.27^{\ddagger}$} & \textcolor{truecolor}{$.84^{\dagger}$ $.42^{\ddagger}$} & \textcolor{truecolor}{$.61^{\dagger}$ $.25^{\ddagger}$} & \textcolor{truecolor}{$.76^{\dagger}$ $.35^{\ddagger}$} \\ &
& FB & \textcolor{fbonly}{$.82^{\dagger}$ $.98^{\ddagger}$} & \textcolor{fbonly}{$.85^{\dagger}$ $.98^{\ddagger}$} & \textcolor{fbonly}{$.99^{\dagger}$ $.99^{\ddagger}$} & \textcolor{fbonly}{$.86^{\dagger}$ $.97^{\ddagger}$}  \\
&          \fontsize{8pt}{8pt}\selectfont \textit{Fwd. Belief} \normalfont               & TB $\land$ FB & \textcolor{falsecolor}{$.65^{\dagger}$ $.25^{\ddagger}$} & \textcolor{falsecolor}{$.69^{\dagger}$ $.34^{\ddagger}$} & \textcolor{falsecolor}{$.60^{\dagger}$ $.24^{\ddagger}$} & \textcolor{falsecolor}{$.63^{\dagger}$ $.32^{\ddagger}$}  \\

\cmidrule{2-7}
\addlinespace[0.5em]

& \raisebox{0.15cm}{\multirow{2}{*}{\includegraphics[width=0.065\textwidth]{figures/table_fwd_action.pdf}}} & TB & \textcolor{truecolor}{$.96^{\dagger}$ $.96^{\ddagger}$} & \textcolor{truecolor}{$.97^{\dagger}$ $.99^{\ddagger}$} & \textcolor{truecolor}{$.99^{\dagger}$ $.99^{\ddagger}$} & \textcolor{truecolor}{$.98^{\dagger}$ $.98^{\ddagger}$}  \\ &
& FB & \textcolor{fbonly}{$.27^{\dagger}$ $.31^{\ddagger}$} & \textcolor{fbonly}{$.22^{\dagger}$ $.30^{\ddagger}$} & \textcolor{fbonly}{$.56^{\dagger}$ $.66^{\ddagger}$} & \textcolor{fbonly}{$.19^{\dagger}$ $.23^{\ddagger}$}  \\
&          \fontsize{8pt}{8pt}\selectfont \textit{Fwd. Action} \normalfont                   & TB $\land$ FB & \textcolor{falsecolor}{$.25^{\dagger}$ $.29^{\ddagger}$} & \textcolor{falsecolor}{$.21^{\dagger}$ $.29^{\ddagger}$}  & \textcolor{falsecolor}{$.56^{\dagger}$ $.65^{\ddagger}$} & \textcolor{falsecolor}{$.18^{\dagger}$ $.22^{\ddagger}$}  \\

\cmidrule{2-7}
\addlinespace[0.5em]

& \raisebox{0.15cm}{\multirow{2}{*}{\includegraphics[width=0.065\textwidth]{figures/table_bwd_belief.pdf}}} & TB & \textcolor{truecolor}{$.54^{\dagger}$ $.12^{\ddagger}$} & \textcolor{truecolor}{$.65^{\dagger}$ $.22^{\ddagger}$} & \textcolor{truecolor}{$.40^{\dagger}$ $.10^{\ddagger}$} & \textcolor{truecolor}{$.54^{\dagger}$ $.19^{\ddagger}$}  \\ &
& FB & \textcolor{fbonly}{$.59^{\dagger}$ $.96^{\ddagger}$} & \textcolor{fbonly}{$.57^{\dagger}$ $.92^{\ddagger}$} & \textcolor{fbonly}{$.84^{\dagger}$ $.97^{\ddagger}$} & \textcolor{fbonly}{$.65^{\dagger}$ $.92^{\ddagger}$}  \\
&          \fontsize{8pt}{8pt}\selectfont \textit{Bwd. Belief} \normalfont             & TB $\land$ FB & \textcolor{falsecolor}{$.24^{\dagger}$ $.09^{\ddagger}$} & \textcolor{falsecolor}{$.30^{\dagger}$ $.16^{\ddagger}$} & \textcolor{falsecolor}{$.28^{\dagger}$ $.07^{\ddagger}$} & \textcolor{falsecolor}{$.25^{\dagger}$ $.14^{\ddagger}$}  \\

\addlinespace[0.5em]
\midrule
\addlinespace[0.5em]
 
\addlinespace[0.5em]
\multirow{11}{*}{\textbf{gpt-3.5}} 

& \raisebox{0.15cm}{\multirow{2}{*}{\includegraphics[width=0.065\textwidth]{figures/table_fwd_belief.pdf}}} & TB & \textcolor{truecolor}{$.81^{\dagger}$ $.35^{\ddagger}$} & \textcolor{truecolor}{$.90^{\dagger}$ $.54^{\ddagger}$} & \textcolor{truecolor}{$.83^{\dagger}$ $.41^{\ddagger}$} & \textcolor{truecolor}{$.95^{\dagger}$ $.97^{\ddagger}$} \\ 
& & FB & \textcolor{fbonly}{$.69^{\dagger}$ $.95^{\ddagger}$} & \textcolor{fbonly}{$.54^{\dagger}$ $.86^{\ddagger}$} & \textcolor{fbonly}{$.88^{\dagger}$ $.97^{\ddagger}$} & \textcolor{fbonly}{$.97^{\dagger}$ $.97^{\ddagger}$}  \\
& \fontsize{8pt}{8pt}\selectfont \textit{Fwd. Belief} \normalfont    & TB $\land$ FB & \textcolor{falsecolor}{$.53^{\dagger}$ $.31^{\ddagger}$} & \textcolor{falsecolor}{$.48^{\dagger}$ $.45^{\ddagger}$} & \textcolor{falsecolor}{$.73^{\dagger}$ $.38^{\ddagger}$} & \textcolor{falsecolor}{$.93^{\dagger}$ $.85^{\ddagger}$}  \\
\cmidrule{2-7}
\addlinespace[0.5em]

& \raisebox{0.15cm}{\multirow{2}{*}{\includegraphics[width=0.065\textwidth]{figures/table_fwd_action.pdf}}} & TB & \textcolor{truecolor}{$.97^{\dagger}$ $.97^{\ddagger}$} & \textcolor{truecolor}{$.96^{\dagger}$ $.97^{\ddagger}$} & \textcolor{truecolor}{$.96^{\dagger}$ $.96^{\ddagger}$} & \textcolor{truecolor}{$.96^{\dagger}$ $.97^{\ddagger}$}  \\ &
& FB & \textcolor{fbonly}{$.19^{\dagger}$ $.22^{\ddagger}$} & \textcolor{fbonly}{$.11^{\dagger}$ $.16^{\ddagger}$} & \textcolor{fbonly}{$.59^{\dagger}$ $.73^{\ddagger}$} & \textcolor{fbonly}{$.72^{\dagger}$ $.83^{\ddagger}$}  \\
&             \fontsize{8pt}{8pt}\selectfont \textit{Fwd. Action} \normalfont               & TB $\land$ FB & \textcolor{falsecolor}{$.17^{\dagger}$ $.20^{\ddagger}$} & \textcolor{falsecolor}{$.10^{\dagger}$ $.14^{\ddagger}$} & \textcolor{falsecolor}{$.55^{\dagger}$ $.69^{\ddagger}$} & \textcolor{falsecolor}{$.69^{\dagger}$ $.81^{\ddagger}$}  \\

\cmidrule{2-7}
\addlinespace[0.5em]

& \raisebox{0.15cm}{\multirow{2}{*}{\includegraphics[width=0.065\textwidth]{figures/table_bwd_belief.pdf}}} & TB & \textcolor{truecolor}{$.55^{\dagger}$ $.10^{\ddagger}$} & \textcolor{truecolor}{$.62^{\dagger}$ $.19^{\ddagger}$} & \textcolor{truecolor}{$.55^{\dagger}$ $.23^{\ddagger}$} & \textcolor{truecolor}{$.86^{\dagger}$ $.70^{\ddagger}$} \\ 
&  & FB & \textcolor{fbonly}{$.45^{\dagger}$ $.92^{\ddagger}$} & \textcolor{fbonly}{$.30^{\dagger}$ $.77^{\ddagger}$} & \textcolor{fbonly}{$.65^{\dagger}$ $.87^{\ddagger}$} & \textcolor{fbonly}{$.51^{\dagger}$ $.76^{\ddagger}$} \\
& {\fontsize{8pt}{8pt}\selectfont \textit{Bwd. Belief} \normalfont}   & TB $\land$ FB & \textcolor{falsecolor}{$.18^{\dagger}$ $.07^{\ddagger}$} & \textcolor{falsecolor}{$.09^{\dagger}$ $.06^{\ddagger}$} & \textcolor{falsecolor}{$.28^{\dagger}$ $.15^{\ddagger}$} & \textcolor{falsecolor}{$.38^{\dagger}$ $.49^{\ddagger}$} \\

\addlinespace[0.5em]
\midrule
\addlinespace[0.5em]

\addlinespace[0.5em]
\multirow{11}{*}{\textbf{claude}} 

& \raisebox{0.15cm}{\multirow{2}{*}{\includegraphics[width=0.065\textwidth]{figures/table_fwd_belief.pdf}}} & TB & \textcolor{truecolor}{$.97^{\dagger}$ $.62^{\ddagger}$} & \textcolor{truecolor}{$.90^{\dagger}$ $.61^{\ddagger}$} & \textcolor{truecolor}{$.92^{\dagger}$ $.82^{\ddagger}$} & \textcolor{truecolor}{$1.00^{\dagger}$ $.98^{\ddagger}$} \\ 
& & FB & \textcolor{fbonly}{$.82^{\dagger}$ $.97^{\ddagger}$} & \textcolor{fbonly}{$.88^{\dagger}$ $.98^{\ddagger}$} & \textcolor{fbonly}{$.98^{\dagger}$ $.99^{\ddagger}$} & \textcolor{fbonly}{$.99^{\dagger}$ $.99^{\ddagger}$} \\
& \fontsize{8pt}{8pt}\selectfont \textit{Fwd. Belief} \normalfont   & TB $\land$ FB & \textcolor{falsecolor}{$.81^{\dagger}$ $.59^{\ddagger}$} & \textcolor{falsecolor}{$.80^{\dagger}$ $.59^{\ddagger}$} & \textcolor{falsecolor}{$.91^{\dagger}$ $.82^{\ddagger}$} & \textcolor{falsecolor}{$.99^{\dagger}$ $.97^{\ddagger}$} \\

\cmidrule{2-7}
\addlinespace[0.5em]
                            
& \raisebox{0.15cm}{\multirow{2}{*}{\includegraphics[width=0.065\textwidth]{figures/table_fwd_action.pdf}}} & TB & \textcolor{truecolor}{$.98^{\dagger}$ $.99^{\ddagger}$} & \textcolor{truecolor}{$.97^{\dagger}$ $.95^{\ddagger}$} & \textcolor{truecolor}{$.96^{\dagger}$ $.96^{\ddagger}$} & \textcolor{truecolor}{$1.00^{\dagger}$ $1.00^{\ddagger}$} \\ 
& & FB & \textcolor{fbonly}{$.28^{\dagger}$ $.43^{\ddagger}$} & \textcolor{fbonly}{$.43^{\dagger}$ $.49^{\ddagger}$} & \textcolor{fbonly}{$.92^{\dagger}$ $.98^{\ddagger}$} & \textcolor{fbonly}{$.98^{\dagger}$ $1.00^{\ddagger}$} \\
&\fontsize{8pt}{8pt}\selectfont \textit{Fwd. Action} \normalfont   & TB $\land$ FB & \textcolor{falsecolor}{$.27^{\dagger}$ $.42^{\ddagger}$} & \textcolor{falsecolor}{$.41^{\dagger}$ $.46^{\ddagger}$} & \textcolor{falsecolor}{$.88^{\dagger}$ $.93^{\ddagger}$} & \textcolor{falsecolor}{$.97^{\dagger}$ $1.00^{\ddagger}$}  \\

\cmidrule{2-7}
\addlinespace[0.5em]

& \raisebox{0.15cm}{\multirow{2}{*}{\includegraphics[width=0.065\textwidth]{figures/table_bwd_belief.pdf}}} & TB & \textcolor{truecolor}{$.79^{\dagger}$ $.29^{\ddagger}$} & \textcolor{truecolor}{$.74^{\dagger}$ $.29^{\ddagger}$} & \textcolor{truecolor}{$.59^{\dagger}$ $.22^{\ddagger}$} & \textcolor{truecolor}{$.79^{\dagger}$ $.73^{\ddagger}$} \\ 
& & FB & \textcolor{fbonly}{$.48^{\dagger}$ $.80^{\ddagger}$} & \textcolor{fbonly}{$.55^{\dagger}$ $.83^{\ddagger}$} & \textcolor{fbonly}{$.89^{\dagger}$ $.96^{\ddagger}$} & \textcolor{fbonly}{$.76^{\dagger}$ $.81^{\ddagger}$} \\
& \fontsize{8pt}{8pt}\selectfont \textit{Bwd. Belief} \normalfont    & TB $\land$ FB & \textcolor{falsecolor}{$.33^{\dagger}$ $.15^{\ddagger}$} & \textcolor{falsecolor}{$.39^{\dagger}$ $.22^{\ddagger}$} & \textcolor{falsecolor}{$.48^{\dagger}$ $.20^{\ddagger}$} & \textcolor{falsecolor}{$.56^{\dagger}$ $.55^{\ddagger}$} \\
     
\addlinespace[0.5em]        
\midrule
\addlinespace[0.5em]

\addlinespace[0.5em]
\multirow{11}{*}{\textbf{claude-2}} 

& \raisebox{0.15cm}{\multirow{2}{*}{\includegraphics[width=0.065\textwidth]{figures/table_fwd_belief.pdf}}} & TB & \textcolor{truecolor}{$.88^{\dagger}$ $.55^{\ddagger}$} & \textcolor{truecolor}{$.93^{\dagger}$ $.87^{\ddagger}$} & \textcolor{truecolor}{$.99^{\dagger}$ $.67^{\ddagger}$} & \textcolor{truecolor}{$1.00^{\dagger}$ $.95^{\ddagger}$} \\ 
& & FB & \textcolor{fbonly}{$.75^{\dagger}$ $.94^{\ddagger}$} & \textcolor{fbonly}{$.95^{\dagger}$ $.99^{\ddagger}$} & \textcolor{fbonly}{$.95^{\dagger}$ $.98^{\ddagger}$} & \textcolor{fbonly}{$.99^{\dagger}$ $.99^{\ddagger}$} \\
& \fontsize{8pt}{8pt}\selectfont \textit{Fwd. Belief} \normalfont   & TB $\land$ FB & \textcolor{falsecolor}{$.68^{\dagger}$ $.52^{\ddagger}$} & \textcolor{falsecolor}{$.89^{\dagger}$ $.61^{\ddagger}$} & \textcolor{falsecolor}{$.84^{\dagger}$ $.67^{\ddagger}$} & \textcolor{falsecolor}{$.99^{\dagger}$ $.95^{\ddagger}$} \\

\cmidrule{2-7}
\addlinespace[0.5em]
                            
& \raisebox{0.15cm}{\multirow{2}{*}{\includegraphics[width=0.065\textwidth]{figures/table_fwd_action.pdf}}} & TB & \textcolor{truecolor}{$.95^{\dagger}$ $.96^{\ddagger}$} & \textcolor{truecolor}{$.96^{\dagger}$ $.95^{\ddagger}$} & \textcolor{truecolor}{$.89^{\dagger}$ $.91^{\ddagger}$} & \textcolor{truecolor}{$.99^{\dagger}$ $1.00^{\ddagger}$} \\ 
& & FB & \textcolor{fbonly}{$.36^{\dagger}$ $.49^{\ddagger}$} & \textcolor{fbonly}{$.51^{\dagger}$ $.70^{\ddagger}$} & \textcolor{fbonly}{$.90^{\dagger}$ $1.00^{\ddagger}$} & \textcolor{fbonly}{$.97^{\dagger}$ $.96^{\ddagger}$} \\
&\fontsize{8pt}{8pt}\selectfont \textit{Fwd. Action} \normalfont   & TB $\land$ FB & \textcolor{falsecolor}{$.34^{\dagger}$ $.47^{\ddagger}$} & \textcolor{falsecolor}{$.49^{\dagger}$ $.65^{\ddagger}$} & \textcolor{falsecolor}{$.83^{\dagger}$ $1.00^{\ddagger}$} & \textcolor{falsecolor}{$.95^{\dagger}$ $.96^{\ddagger}$}  \\

\cmidrule{2-7}
\addlinespace[0.5em]

& \raisebox{0.15cm}{\multirow{2}{*}{\includegraphics[width=0.065\textwidth]{figures/table_bwd_belief.pdf}}} & TB & \textcolor{truecolor}{$.75^{\dagger}$ $.30^{\ddagger}$} & \textcolor{truecolor}{$.82^{\dagger}$ $.34^{\ddagger}$} & \textcolor{truecolor}{$.45^{\dagger}$ $.22^{\ddagger}$} & \textcolor{truecolor}{$.61^{\dagger}$ $.39^{\ddagger}$} \\ 
& & FB & \textcolor{fbonly}{$.50^{\dagger}$ $.82^{\ddagger}$} & \textcolor{fbonly}{$.60^{\dagger}$ $.88^{\ddagger}$} & \textcolor{fbonly}{$.89^{\dagger}$ $.96^{\ddagger}$} & \textcolor{fbonly}{$.92^{\dagger}$ $.96^{\ddagger}$} \\
& \fontsize{8pt}{8pt}\selectfont \textit{Bwd. Belief} \normalfont    & TB $\land$ FB & \textcolor{falsecolor}{$.39^{\dagger}$ $.24^{\ddagger}$} & \textcolor{falsecolor}{$.43^{\dagger}$ $.26^{\ddagger}$} & \textcolor{falsecolor}{$.41^{\dagger}$ $.21^{\ddagger}$} & \textcolor{falsecolor}{$.53^{\dagger}$ $.35^{\ddagger}$} \\
     
\addlinespace[0.5em]        
\midrule
\addlinespace[0.5em]
     
\addlinespace[0.5em]
\multirow{11}{*}{\textbf{gpt-4}} 

& \raisebox{0.15cm}{\multirow{2}{*}{\includegraphics[width=0.065\textwidth]{figures/table_fwd_belief.pdf}}} & TB & \textcolor{truecolor}{$.99^{\dagger}$ $.91^{\ddagger}$} & \textcolor{truecolor}{$.99^{\dagger}$ $.99^{\ddagger}$} & \textcolor{truecolor}{$.99^{\dagger}$ $.97^{\ddagger}$} & \textcolor{truecolor}{$1.00^{\dagger}$ $.97^{\ddagger}$} \\ 
& & FB & \textcolor{fbonly}{$.98^{\dagger}$ $.99^{\ddagger}$} & \textcolor{fbonly}{$.99^{\dagger}$ $.99^{\ddagger}$} & \textcolor{fbonly}{$.99^{\dagger}$ $.99^{\ddagger}$} & \textcolor{fbonly}{$.99^{\dagger}$ $.99^{\ddagger}$}  \\
& \fontsize{8pt}{8pt}\selectfont \textit{Fwd. Belief} \normalfont   & TB $\land$ FB & \textcolor{falsecolor}{$.97^{\dagger}$ $.90^{\ddagger}$} & \textcolor{falsecolor}{$.98^{\dagger}$ $.98^{\ddagger}$} & \textcolor{falsecolor}{$.97^{\dagger}$ $.96^{\ddagger}$} & \textcolor{falsecolor}{$.99^{\dagger}$ $.96^{\ddagger}$}  \\

\cmidrule{2-7}
\addlinespace[0.5em]

& \raisebox{0.15cm}{\multirow{2}{*}{\includegraphics[width=0.065\textwidth]{figures/table_fwd_action.pdf}}} & TB & \textcolor{truecolor}{$.98^{\dagger}$ $.98^{\ddagger}$} & \textcolor{truecolor}{$.99^{\dagger}$ $.99^{\ddagger}$} & \textcolor{truecolor}{$1.00^{\dagger}$ $1.00^{\ddagger}$} & \textcolor{truecolor}{$1.00^{\dagger}$ $1.00^{\ddagger}$}  \\ &
& FB & \textcolor{fbonly}{$.81^{\dagger}$ $.92^{\ddagger}$} & \textcolor{fbonly}{$.88^{\dagger}$ $.96^{\ddagger}$} & \textcolor{fbonly}{$.98^{\dagger}$ $1.00^{\ddagger}$} & \textcolor{fbonly}{$1.00^{\dagger}$ $.99^{\ddagger}$}  \\
&          \fontsize{8pt}{8pt}\selectfont \textit{Fwd. Action} \normalfont               & TB $\land$ FB & \textcolor{falsecolor}{$.79^{\dagger}$ $.90^{\ddagger}$} & \textcolor{falsecolor}{$.87^{\dagger}$ $.95^{\ddagger}$} & \textcolor{falsecolor}{$.98^{\dagger}$ $1.00^{\ddagger}$} & \textcolor{falsecolor}{$1.00^{\dagger}$ $.99^{\ddagger}$}  \\

\cmidrule{2-7}
\addlinespace[0.5em]

& \raisebox{0.15cm}{\multirow{2}{*}{\includegraphics[width=0.065\textwidth]{figures/table_bwd_belief.pdf}}} & TB & \textcolor{truecolor}{$.86^{\dagger}$ $.62^{\ddagger}$} & \textcolor{truecolor}{$.84^{\dagger}$ $.76^{\ddagger}$} & \textcolor{truecolor}{$.68^{\dagger}$ $.57^{\ddagger}$} & \textcolor{truecolor}{$.83^{\dagger}$ $.81^{\ddagger}$}  \\ &
& FB & \textcolor{fbonly}{$.53^{\dagger}$ $.77^{\ddagger}$} & \textcolor{fbonly}{$.54^{\dagger}$ $.63^{\dagger}$} & \textcolor{fbonly}{$.85^{\dagger}$ $.92^{\ddagger}$} & \textcolor{fbonly}{$.75^{\dagger}$ $.85^{\ddagger}$}  \\
&          \fontsize{8pt}{8pt}\selectfont \textit{Bwd. Belief \normalfont}               & TB $\land$ FB & \textcolor{falsecolor}{$.40^{\dagger}$ $.40^{\ddagger}$} & \textcolor{falsecolor}{$.38^{\dagger}$ $.40^{\ddagger}$} & \textcolor{falsecolor}{$.53^{\dagger}$ $.49^{\ddagger}$} & \textcolor{falsecolor}{$.58^{\dagger}$ $.65^{\ddagger}$}  \\
                           
\addlinespace[0.5em]
    
    \bottomrule
  \end{tabular}
\end{table}

\pagebreak


\begin{table}
\vspace{-1cm}
  \caption{Model performance for controls. TB = \textcolor{truecolor}{True Belief}. FB = \textcolor{falsecolor}{False Belief}. $^{\dagger}$ = without initial belief. $^{\ddagger}$ = with initial belief.}
  \label{tab:arescontrol}
  \centering
  \fontsize{8.5pt}{8.5pt}\selectfont
  \begin{tabular}{lccc}
    \toprule
    \multirow{2}{*}{\textbf{Model}} & \multirow{2}{*}{\textbf{Condition}} & \multirow{2}{*}{\textbf{Contingency}} & \multicolumn{1}{c}{\textbf{Method}}  \\
  \cline{4-4} 
  \addlinespace[0.5em]
   & & & \textit{0-shot} \\
  \midrule
  \addlinespace[0.5em]
  \addlinespace[0.5em]
  \multirow{11}{*}{\textbf{llama-65}} & 
  
  \raisebox{0.15cm}{\multirow{2}{*}{\includegraphics[width=0.065\textwidth]{figures/table_fwd_belief.pdf}}} & TB & \textcolor{truecolor}{.81$^{\dagger}$ .94$^{\ddagger}$} \\
  & & FB & \textcolor{fbonly}{.86$^{\dagger}$ .96$^{\ddagger}$}\\
  &  \fontsize{8pt}{8pt}\selectfont \textit{Fwd. Belief} \normalfont & TB $\land$ FB & \textcolor{falsecolor}{.79$^{\dagger}$ .92$^{\dagger}$} \\
  
  \addlinespace[0.5em]
  \cmidrule{2-4}
  \addlinespace[0.5em]

  & \raisebox{0.15cm}{\multirow{2}{*}{\includegraphics[width=0.065\textwidth]{figures/table_fwd_action.pdf}}} & TB & \textcolor{truecolor}{.53$^{\dagger}$ .68$^{\ddagger}$} \\
  & & FB & \textcolor{fbonly}{.56$^{\dagger}$ .72$^{\ddagger}$}\\
  &  \fontsize{8pt}{8pt}\selectfont \textit{Fwd. Action} \normalfont & TB $\land$ FB & \textcolor{falsecolor}{.50$^{\dagger}$ .66$^{\dagger}$} \\

   \addlinespace[0.5em]
  \cmidrule{2-4}
  \addlinespace[0.5em]

  & \raisebox{0.15cm}{\multirow{2}{*}{\includegraphics[width=0.065\textwidth]{figures/table_bwd_belief.pdf}}} & TB & \textcolor{truecolor}{.90$^{\dagger}$ .96$^{\ddagger}$} \\
  & & FB & \textcolor{fbonly}{.90$^{\dagger}$ .96$^{\ddagger}$}\\
  &  \fontsize{8pt}{8pt}\selectfont \textit{Bwd. Belief} \normalfont & TB $\land$ FB & \textcolor{falsecolor}{.90$^{\dagger}$ .96$^{\dagger}$} \\

  \addlinespace[0.5em]
   \midrule
   
  \addlinespace[0.5em]

  \multirow{11}{*}{\textbf{dav-003}} & 
  
  \raisebox{0.15cm}{\multirow{2}{*}{\includegraphics[width=0.065\textwidth]{figures/table_fwd_belief.pdf}}} & TB & \textcolor{truecolor}{.98$^{\dagger}$ .99$^{\ddagger}$} \\
  & & FB & \textcolor{fbonly}{.98$^{\dagger}$ .99$^{\ddagger}$}\\
  &  \fontsize{8pt}{8pt}\selectfont \textit{Fwd. Belief} \normalfont & TB $\land$ FB & \textcolor{falsecolor}{.98$^{\dagger}$ .99$^{\dagger}$} \\
  
  \addlinespace[0.5em]
  \cmidrule{2-4}
  \addlinespace[0.5em]

  & \raisebox{0.15cm}{\multirow{2}{*}{\includegraphics[width=0.065\textwidth]{figures/table_fwd_action.pdf}}} & TB & \textcolor{truecolor}{.88$^{\dagger}$ .91$^{\ddagger}$} \\
  & & FB & \textcolor{fbonly}{.93$^{\dagger}$ .94$^{\ddagger}$}\\
  &  \fontsize{8pt}{8pt}\selectfont \textit{Fwd. Action} \normalfont & TB $\land$ FB & \textcolor{falsecolor}{.88$^{\dagger}$ .90$^{\dagger}$} \\

   \addlinespace[0.5em]
  \cmidrule{2-4}
  \addlinespace[0.5em]

  & \raisebox{0.15cm}{\multirow{2}{*}{\includegraphics[width=0.065\textwidth]{figures/table_bwd_belief.pdf}}} & TB & \textcolor{truecolor}{.98$^{\dagger}$ .99$^{\ddagger}$} \\
  & & FB & \textcolor{fbonly}{.98$^{\dagger}$ .99$^{\ddagger}$}\\
  &  \fontsize{8pt}{8pt}\selectfont \textit{Bwd. Belief} \normalfont & TB $\land$ FB & \textcolor{falsecolor}{.98$^{\dagger}$ .99$^{\dagger}$} \\

      \addlinespace[0.5em]
   \midrule
   
  \addlinespace[0.5em]

  \multirow{11}{*}{\textbf{gpt-3.5}} & 
  
  \raisebox{0.15cm}{\multirow{2}{*}{\includegraphics[width=0.065\textwidth]{figures/table_fwd_belief.pdf}}} & TB & \textcolor{truecolor}{.99$^{\dagger}$ .99$^{\ddagger}$} \\
  & & FB & \textcolor{fbonly}{.98$^{\dagger}$ .99$^{\ddagger}$}\\
  &  \fontsize{8pt}{8pt}\selectfont \textit{Fwd. Belief} \normalfont & TB $\land$ FB & \textcolor{falsecolor}{.98$^{\dagger}$ .99$^{\dagger}$} \\
  
  \addlinespace[0.5em]
  \cmidrule{2-4}
  \addlinespace[0.5em]

  & \raisebox{0.15cm}{\multirow{2}{*}{\includegraphics[width=0.065\textwidth]{figures/table_fwd_action.pdf}}} & TB & \textcolor{truecolor}{.83$^{\dagger}$ .85$^{\ddagger}$} \\
  & & FB & \textcolor{fbonly}{.82$^{\dagger}$ .87$^{\ddagger}$}\\
  &  \fontsize{8pt}{8pt}\selectfont \textit{Fwd. Action} \normalfont & TB $\land$ FB & \textcolor{falsecolor}{.77$^{\dagger}$ .79$^{\dagger}$} \\

   \addlinespace[0.5em]
  \cmidrule{2-4}
  \addlinespace[0.5em]

  & \raisebox{0.15cm}{\multirow{2}{*}{\includegraphics[width=0.065\textwidth]{figures/table_bwd_belief.pdf}}} & TB & \textcolor{truecolor}{.99$^{\dagger}$ .99$^{\ddagger}$} \\
  & & FB & \textcolor{fbonly}{.99$^{\dagger}$ .99$^{\ddagger}$}\\
  &  \fontsize{8pt}{8pt}\selectfont \textit{Bwd. Belief} \normalfont & TB $\land$ FB & \textcolor{falsecolor}{.99$^{\dagger}$ .99$^{\dagger}$} \\

 \addlinespace[0.5em]
   \midrule
   
  \addlinespace[0.5em]

  \multirow{11}{*}{\textbf{claude}} & 
  
  \raisebox{0.15cm}{\multirow{2}{*}{\includegraphics[width=0.065\textwidth]{figures/table_fwd_belief.pdf}}} & TB & \textcolor{truecolor}{.99$^{\dagger}$ .99$^{\ddagger}$} \\
  & & FB & \textcolor{fbonly}{.99$^{\dagger}$ .99$^{\ddagger}$}\\
  &  \fontsize{8pt}{8pt}\selectfont \textit{Fwd. Belief} \normalfont & TB $\land$ FB & \textcolor{falsecolor}{.99$^{\dagger}$ .99$^{\dagger}$} \\
  
  \addlinespace[0.5em]
  \cmidrule{2-4}
  \addlinespace[0.5em]

  & \raisebox{0.15cm}{\multirow{2}{*}{\includegraphics[width=0.065\textwidth]{figures/table_fwd_action.pdf}}} & TB & \textcolor{truecolor}{.88$^{\dagger}$ .93$^{\ddagger}$} \\
  & & FB & \textcolor{fbonly}{.86$^{\dagger}$ .93$^{\ddagger}$}\\
  &  \fontsize{8pt}{8pt}\selectfont \textit{Fwd. Action} \normalfont & TB $\land$ FB & \textcolor{falsecolor}{.83$^{\dagger}$ .89$^{\dagger}$} \\

   \addlinespace[0.5em]
  \cmidrule{2-4}
  \addlinespace[0.5em]

  & \raisebox{0.15cm}{\multirow{2}{*}{\includegraphics[width=0.065\textwidth]{figures/table_bwd_belief.pdf}}} & TB & \textcolor{truecolor}{.99$^{\dagger}$ .99$^{\ddagger}$} \\
  & & FB & \textcolor{fbonly}{.99$^{\dagger}$ .99$^{\ddagger}$}\\
  &  \fontsize{8pt}{8pt}\selectfont \textit{Bwd. Belief} \normalfont & TB $\land$ FB & \textcolor{falsecolor}{.99$^{\dagger}$ .99$^{\dagger}$} \\

 \addlinespace[0.5em]
   \midrule
   
  \addlinespace[0.5em]

  \multirow{11}{*}{\textbf{gpt-4}} & 
  
  \raisebox{0.15cm}{\multirow{2}{*}{\includegraphics[width=0.065\textwidth]{figures/table_fwd_belief.pdf}}} & TB & \textcolor{truecolor}{.99$^{\dagger}$ .99$^{\ddagger}$} \\
  & & FB & \textcolor{fbonly}{.99$^{\dagger}$ .99$^{\ddagger}$}\\
  &  \fontsize{8pt}{8pt}\selectfont \textit{Fwd. Belief} \normalfont & TB $\land$ FB & \textcolor{falsecolor}{.99$^{\dagger}$ .99$^{\dagger}$} \\
  
  \addlinespace[0.5em]
  \cmidrule{2-4}
  \addlinespace[0.5em]

  & \raisebox{0.15cm}{\multirow{2}{*}{\includegraphics[width=0.065\textwidth]{figures/table_fwd_action.pdf}}} & TB & \textcolor{truecolor}{.98$^{\dagger}$ .95$^{\ddagger}$} \\
  & & FB & \textcolor{fbonly}{.99$^{\dagger}$ .98$^{\ddagger}$}\\
  &  \fontsize{8pt}{8pt}\selectfont \textit{Fwd. Action} \normalfont & TB $\land$ FB & \textcolor{falsecolor}{.98$^{\dagger}$ .94$^{\dagger}$} \\

   \addlinespace[0.5em]
  \cmidrule{2-4}
  \addlinespace[0.5em]

  & \raisebox{0.15cm}{\multirow{2}{*}{\includegraphics[width=0.065\textwidth]{figures/table_bwd_belief.pdf}}} & TB & \textcolor{truecolor}{.99$^{\dagger}$ .99$^{\ddagger}$} \\
  & & FB & \textcolor{fbonly}{.99$^{\dagger}$ .99$^{\ddagger}$}\\
  &  \fontsize{8pt}{8pt}\selectfont \textit{Bwd. Belief} \normalfont & TB $\land$ FB & \textcolor{falsecolor}{.99$^{\dagger}$ .99$^{\dagger}$} \\

  \bottomrule
  \end{tabular}
\end{table}

\pagebreak

\begin{table}
  \caption{Model Performance Initial Percept to Initial Belief.}
  \label{tab:arescontrolpb}
  \centering
  \fontsize{8.5pt}{8.5pt}\selectfont
  \begin{tabular}{lc}

    \toprule
    \textbf{Model} & \multicolumn{1}{c}{\textbf{Method}} \\
   \cline{2-2} 
  \addlinespace[0.5em]
   & \textit{0-shot} \\
  \midrule
  \addlinespace[0.5em]

    \textbf{llama-65} & .92\\
    \textbf{dav-003} & .98\\
    \textbf{gpt-3.5} & .96\\
    \textbf{claude} & .98\\
    \textbf{gpt-4} & .99\\
  \bottomrule
  \end{tabular}
\end{table}

\section{Addressing Concerns of Circularity}
\label{asec:circ}

We generate an alternate evaluation set using Claude-2 with 30 populated templates and 750 conditions called BigToM-Claude. We evaluate the source models, Claude-2 and gpt-4 on this evaluation set. The performance of gpt-4-0314 on BigToM-Claude is similar to that of BigToM-gpt4 (see \autoref{tab:ares-claude}). This shows that our tests are consistent regardless of the source that populates the template. Moreover, Claude-2, despite generating BigToM-Claude performs worse than gpt-4 on this dataset. With this validation set generated by Claude, we are able to show that the abilities tested during evaluation are distinct from the ones that are used while populating the template. To further validate this, we ran a logistic regression with data source (BigToM-gpt4, BigToM-Claude) and model type (gpt-4, Claude-2) as predictors and response (correct, incorrect) as dependent variable. To match the smaller size of BigToM-Claude, we randomly sampled a subset of BigToM-gpt4. The results of this analysis confirmed that there was no significant interaction ($b$ = -0.13, $p$ = 0.620), while there was a significant main effect of model type, with gpt-4 performing better across conditions than Claude-2 ($b$ = 1.07, $p$ < .001). Model performances for each data source are shown in \autoref{tab:aclaudevsgpt}.

\begin{table}
  \caption{Claude-2 vs. gpt-4 accuracy on BigTom-Claude and BigTom-gpt4.} 
  \label{tab:aclaudevsgpt}
  \centering
  \fontsize{8.5pt}{8.5pt}\selectfont
  \begin{tabular}{cccc}
    \toprule
    & \multicolumn{2}{c}{\textbf{Data Source}} & \\
    \cline{2-3}
    \addlinespace[0.5em]
    \multirow{1}{*}{\textbf{Model Name}} & \textit{BigTom-Claude} & \textit{BigTom-gpt4} \\
  
    \midrule
    \addlinespace[0.5em]
    Claude-2 & $0.675$  & $0.708$ \\ 
    \addlinespace[0.5em]
    gpt-4 & $0.858$  & $0.861$ \\ 

    \bottomrule
  \end{tabular}
\end{table}

\begin{table}
  \caption{GPT-4's and Claude-2's performance for each method on 30 test items generated by Claude-2. TB = \textcolor{truecolor}{True Belief}. FB = \textcolor{falsecolor}{False Belief}. $^{\dagger}$ = without initial belief. $^{\ddagger}$ = with initial belief.}
  \label{tab:ares-claude}
  \centering
  \fontsize{8.5pt}{8.5pt}\selectfont
  \begin{tabular}{lcccccc}
    \toprule
    \multirow{2}{*}{\textbf{Model}} & \multirow{2}{*}{\textbf{Condition}} & \multirow{2}{*}{\textbf{Contingency}} & \multicolumn{4}{c}{\textbf{Method}} \\
  \cline{4-7} 
  \addlinespace[0.5em]
  
  & & & \textit{0-shot} & \textit{0-shot-cot} & \textit{1-shot} & \textit{1-shot-cot} \\
  
\midrule
\addlinespace[0.5em]

\addlinespace[0.5em]
\multirow{11}{*}{\textbf{claude-2}} 

& \raisebox{0.15cm}{\multirow{2}{*}{\includegraphics[width=0.065\textwidth]{figures/table_fwd_belief.pdf}}} & TB & \textcolor{truecolor}{$1.00^{\dagger}$ $.60^{\ddagger}$} & \textcolor{truecolor}{$.97^{\dagger}$ $.80^{\ddagger}$} & \textcolor{truecolor}{$.60^{\dagger}$ $.34^{\ddagger}$} & \textcolor{truecolor}{$.90^{\dagger}$ $.77^{\ddagger}$} \\ 
& & FB & \textcolor{fbonly}{$.67^{\dagger}.97^{\ddagger}$} & \textcolor{fbonly}{$.94^{\dagger}$ $.97^{\ddagger}$} & \textcolor{fbonly}{$.97^{\dagger}$ $1.00^{\ddagger}$} & \textcolor{fbonly}{$1.00^{\dagger}$ $1.00^{\ddagger}$} \\
& \fontsize{8pt}{8pt}\selectfont \textit{Fwd. Belief} \normalfont   & TB $\land$ FB & \textcolor{falsecolor}{$.67^{\dagger}$ $.57^{\ddagger}$} & \textcolor{falsecolor}{$.90^{\dagger}$ $.77^{\ddagger}$} & \textcolor{falsecolor}{$.60^{\dagger}$ $.34^{\ddagger}$} & \textcolor{falsecolor}{$.90^{\dagger}$ $.77^{\ddagger}$} \\

\cmidrule{2-7}
\addlinespace[0.5em]
                            
& \raisebox{0.15cm}{\multirow{2}{*}{\includegraphics[width=0.065\textwidth]{figures/table_fwd_action.pdf}}} & TB & \textcolor{truecolor}{$.94^{\dagger}$ $.94^{\ddagger}$} & \textcolor{truecolor}{$.90^{\dagger}$ $.97^{\ddagger}$} & \textcolor{truecolor}{$.84^{\dagger}$ $.80^{\ddagger}$} & \textcolor{truecolor}{$.97^{\dagger}$ $1.00^{\ddagger}$} \\ 
& & FB & \textcolor{fbonly}{$.34^{\dagger}$ $.54^{\ddagger}$} & \textcolor{fbonly}{$.54^{\dagger}$ $.78^{\ddagger}$} & \textcolor{fbonly}{$.84^{\dagger}$ $.77^{\ddagger}$} & \textcolor{fbonly}{$.97^{\dagger}$ $1.00^{\ddagger}$} \\
&\fontsize{8pt}{8pt}\selectfont \textit{Fwd. Action} \normalfont   & TB $\land$ FB & \textcolor{falsecolor}{$.30^{\dagger}$ $.47^{\ddagger}$} & \textcolor{falsecolor}{$.47^{\dagger}$ $.74^{\ddagger}$} & \textcolor{falsecolor}{$.74^{\dagger}$ $.64^{\ddagger}$} & \textcolor{falsecolor}{$.94^{\dagger}$ $1.00^{\ddagger}$}  \\

\cmidrule{2-7}
\addlinespace[0.5em]

& \raisebox{0.15cm}{\multirow{2}{*}{\includegraphics[width=0.065\textwidth]{figures/table_bwd_belief.pdf}}} & TB & \textcolor{truecolor}{$.84^{\dagger}$ $.40^{\ddagger}$} & \textcolor{truecolor}{$.97^{\dagger}$ $.54^{\ddagger}$} & \textcolor{truecolor}{$.37^{\dagger}$ $.30^{\ddagger}$} & \textcolor{truecolor}{$.70^{\dagger}$ $.50^{\ddagger}$} \\ 
& & FB & \textcolor{fbonly}{$.17^{\dagger}$ $.74^{\ddagger}$} & \textcolor{fbonly}{$.37^{\dagger}$ $.84^{\ddagger}$} & \textcolor{fbonly}{$.77^{\dagger}$ $.87^{\ddagger}$} & \textcolor{fbonly}{$.97^{\dagger}$ $1.00^{\ddagger}$} \\
& \fontsize{8pt}{8pt}\selectfont \textit{Bwd. Belief} \normalfont    & TB $\land$ FB & \textcolor{falsecolor}{$.10^{\dagger}$ $.24^{\ddagger}$} & \textcolor{falsecolor}{$.34^{\dagger}$ $.47^{\ddagger}$} & \textcolor{falsecolor}{$.34^{\dagger}$ $.30^{\ddagger}$} & \textcolor{falsecolor}{$.70^{\dagger}$ $.50^{\ddagger}$} \\
     
\addlinespace[0.5em]        
\midrule
\addlinespace[0.5em]
     
\addlinespace[0.5em]
\multirow{11}{*}{\textbf{gpt-4}} 

& \raisebox{0.15cm}{\multirow{2}{*}{\includegraphics[width=0.065\textwidth]{figures/table_fwd_belief.pdf}}} & TB & \textcolor{truecolor}{$.97^{\dagger}$ $.94^{\ddagger}$} & \textcolor{truecolor}{$.97^{\dagger}$ $.87^{\ddagger}$} & \textcolor{truecolor}{$.97^{\dagger}$ $.94^{\ddagger}$} & \textcolor{truecolor}{$.97^{\dagger}$ $.94^{\ddagger}$} \\ 
& & FB & \textcolor{fbonly}{$1.00^{\dagger}$ $1.00^{\ddagger}$} & \textcolor{fbonly}{$1.00^{\dagger}$ $1.00^{\ddagger}$} & \textcolor{fbonly}{$1.00^{\dagger}$ $1.00^{\ddagger}$} & \textcolor{fbonly}{$1.00^{\dagger}$ $1.00^{\ddagger}$}  \\
& \fontsize{8pt}{8pt}\selectfont \textit{Fwd. Belief} \normalfont   & TB $\land$ FB & \textcolor{falsecolor}{$.97^{\dagger}$ $.94^{\ddagger}$} & \textcolor{falsecolor}{$.97^{\dagger}$ $.87^{\ddagger}$} & \textcolor{falsecolor}{$.97^{\dagger}$ $.94^{\ddagger}$} & \textcolor{falsecolor}{$.97^{\dagger}$ $.94^{\ddagger}$}  \\

\cmidrule{2-7}
\addlinespace[0.5em]

& \raisebox{0.15cm}{\multirow{2}{*}{\includegraphics[width=0.065\textwidth]{figures/table_fwd_action.pdf}}} & TB & \textcolor{truecolor}{$.94^{\dagger}$ $.97^{\ddagger}$} & \textcolor{truecolor}{$.97^{\dagger}$ $.97^{\ddagger}$} & \textcolor{truecolor}{$.94^{\dagger}$ $.97^{\ddagger}$} & \textcolor{truecolor}{$.94^{\dagger}$ $1.00^{\ddagger}$}  \\ &
& FB & \textcolor{fbonly}{$.64^{\dagger}$ $.97^{\ddagger}$} & \textcolor{fbonly}{$.87^{\dagger}$ $.97^{\ddagger}$} & \textcolor{fbonly}{$1.00^{\dagger}$ $1.00^{\ddagger}$} & \textcolor{fbonly}{$1.00^{\dagger}$ $1.00^{\ddagger}$}  \\
&          \fontsize{8pt}{8pt}\selectfont \textit{Fwd. Action} \normalfont               & TB $\land$ FB & \textcolor{falsecolor}{$.57^{\dagger}$ $.94^{\ddagger}$} & \textcolor{falsecolor}{$.84^{\dagger}$ $.94^{\ddagger}$} & \textcolor{falsecolor}{$.94^{\dagger}$ $.97^{\ddagger}$} & \textcolor{falsecolor}{$.94^{\dagger}$ $1.00^{\ddagger}$}  \\

\cmidrule{2-7}
\addlinespace[0.5em]

& \raisebox{0.15cm}{\multirow{2}{*}{\includegraphics[width=0.065\textwidth]{figures/table_bwd_belief.pdf}}} & TB & \textcolor{truecolor}{$1.00^{\dagger}$ $.80^{\ddagger}$} & \textcolor{truecolor}{$.94^{\dagger}$ $.84^{\ddagger}$} & \textcolor{truecolor}{$.84^{\dagger}$ $.70^{\ddagger}$} & \textcolor{truecolor}{$.90^{\dagger}$ $.74^{\ddagger}$}  \\ &
& FB & \textcolor{fbonly}{$.37^{\dagger}$ $.74^{\ddagger}$} & \textcolor{fbonly}{$.50^{\dagger}$ $.70^{\dagger}$} & \textcolor{fbonly}{$.97^{\dagger}$ $1.00^{\ddagger}$} & \textcolor{fbonly}{$.97^{\dagger}$ $.97^{\ddagger}$}  \\
&          \fontsize{8pt}{8pt}\selectfont \textit{Bwd. Belief \normalfont}               & TB $\land$ FB & \textcolor{falsecolor}{$.37^{\dagger}$ $.54^{\ddagger}$} & \textcolor{falsecolor}{$.44^{\dagger}$ $.54^{\ddagger}$} & \textcolor{falsecolor}{$.84^{\dagger}$ $.70^{\ddagger}$} & \textcolor{falsecolor}{$.87^{\dagger}$ $.70^{\ddagger}$}  \\
                           
\addlinespace[0.5em]
    
    \bottomrule
  \end{tabular}
\end{table}

\section{Diversity in BigToM}

We analyze the diversity of the dataset by asking GPT-4 to annotate the 200 templates along 3 different dimensions, types of changeable states, types of mechanisms of change, types of medium of perceptions. These are generic labels provided by GPT-4 for each of the stories (see \autoref{tab:div}).

We show some examples of these categories here:

\begin{itemize}
    \item Examples of mechanisms of change: rainfall, monkeys, power surge, leak, toddler, fog, cat, vandals, malfunction
    \item Examples of mediums of perception: tastes, sees, feels, hears, smells, reads, discovers/notices (ambiguous) 
    \item Examples of changeable states: water level, type of ingredient, temperature, visibility, dryness, cleanliness, color, tuning, sharpness
\end{itemize}

These descriptive statistics show that our generated items have substantially diverse content. In contrast, expert written datasets are very limited in their number and hence diversity due to the time and effort involved in generating the evaluations. For example most expert items only involve a change in location of the object (changeable state: location), which is usually done by an agent (mechanism of change: person) and is perceived by looking at the agent changing the object. 

\begin{table}[!htbp]
    \centering
    \caption{Measures of diversity of the evaluations in our dataset.}
      \vspace{0.2cm}
    \begin{tabular}{ccc}
\toprule
Changeable States & Mechanisms of Change & Mediums of perception \\
\midrule
154 & 76 & 7 \\
\bottomrule
    \end{tabular}
\label{tab:div}
\end{table}


\section{Participant Instructions and Pay}
\label{asec:irb}

\begin{figure}
\centering
\begin{tcolorbox}[
prompt,
title={\small \textbf{Instructions Experiment 1.}},
width=0.9\textwidth]
\fontsize{7pt}{7pt}\selectfont
Please read these instructions carefully, as understanding them is important for successfully completing the study. After reading the instructions, you will need to complete a comprehension quiz to confirm that you understood the task.\\

Instructions:\\

In this online experiment, your task is to evaluate the quality of different short social reasoning scenarios. Each scenario consists of a story, a question, and different answer options. We are interested in whether the story is easy to understand and whether the question and answers make sense with respect to the story. We also want to know whether the 'correct' answer to the question is unclear or ambiguous.\\

An example scenario can look like this:\\

\#\#\#\#\\

\{example 1\}\\

\#\#\#\#\\

Another example of a short story can look like this:\\

\#\#\#\#\\

\{example 2\}\\

\#\#\#\#\\

In total, you will evaluate 30 scenarios. It is important to read each of them carefully and respond to the best of your ability.\\
\normalfont
\end{tcolorbox}
\caption{
Instructions shown to participants in Experiment 1.
\label{fig:afig_instructions_exp_1}
}
\end{figure}

\begin{figure}
\centering
\begin{tcolorbox}[
prompt,
title={\small \textbf{Instructions Experiment 2.}},
width=0.9\textwidth]
\fontsize{7pt}{7pt}\selectfont
Please read these instructions carefully, as understanding them is important for successfully completing the study. After reading the instructions, you will need to complete a comprehension quiz to confirm that you understood the task.\\

Instructions:\\

In this online experiment, you will be presented with short social reasoning scenarios. Each scenario includes a story, a question, and answer options. Your task is to select the best answer based on your understanding of the scenario.\\

An example scenario can look like this:\\

\#\#\#\#\\

\{example 1\}\\

\#\#\#\#\\

In total, you will answer 40 questions. It is important to read each story carefully and respond to the questions to the best of your ability.\\
\normalfont
\end{tcolorbox}
\caption{
Instructions shown to participants in Experiment 2.
\label{fig:afig_instructions_exp_2}
}
\end{figure}

\textbf{Personally Identifiable Info and IRB}.
Participants completed a consent page before the start of each experiment. Participants were made aware of potential risks and links to the Stanford IRB approvals were provided on the consent page.

\textbf{Duration and Pay}. For Experiment 1, participants were paid \$$12.02$/hr. The median completion time of Experiment 1 was $21$ minutes and $55$ seconds. The total cost for Experiment 1 was \$$512.00$. For Experiment 2, participants were paid \$$12.05$/hr. The median completion time of Experiment 2 was $21$ minutes and $22$ seconds. The total cost for Experiment 2 was \$$114.40$.


\section{Licenses}
\label{asec:license}
We release the dataset with an MIT License, see \href{https://sites.google.com/view/social-reasoning-lms}{https://sites.google.com/view/social-reasoning-lms}.

\section{Extension to Second Order Beliefs}

\begin{figure}[!htbp]
    \centering
    \includegraphics[width=\textwidth]{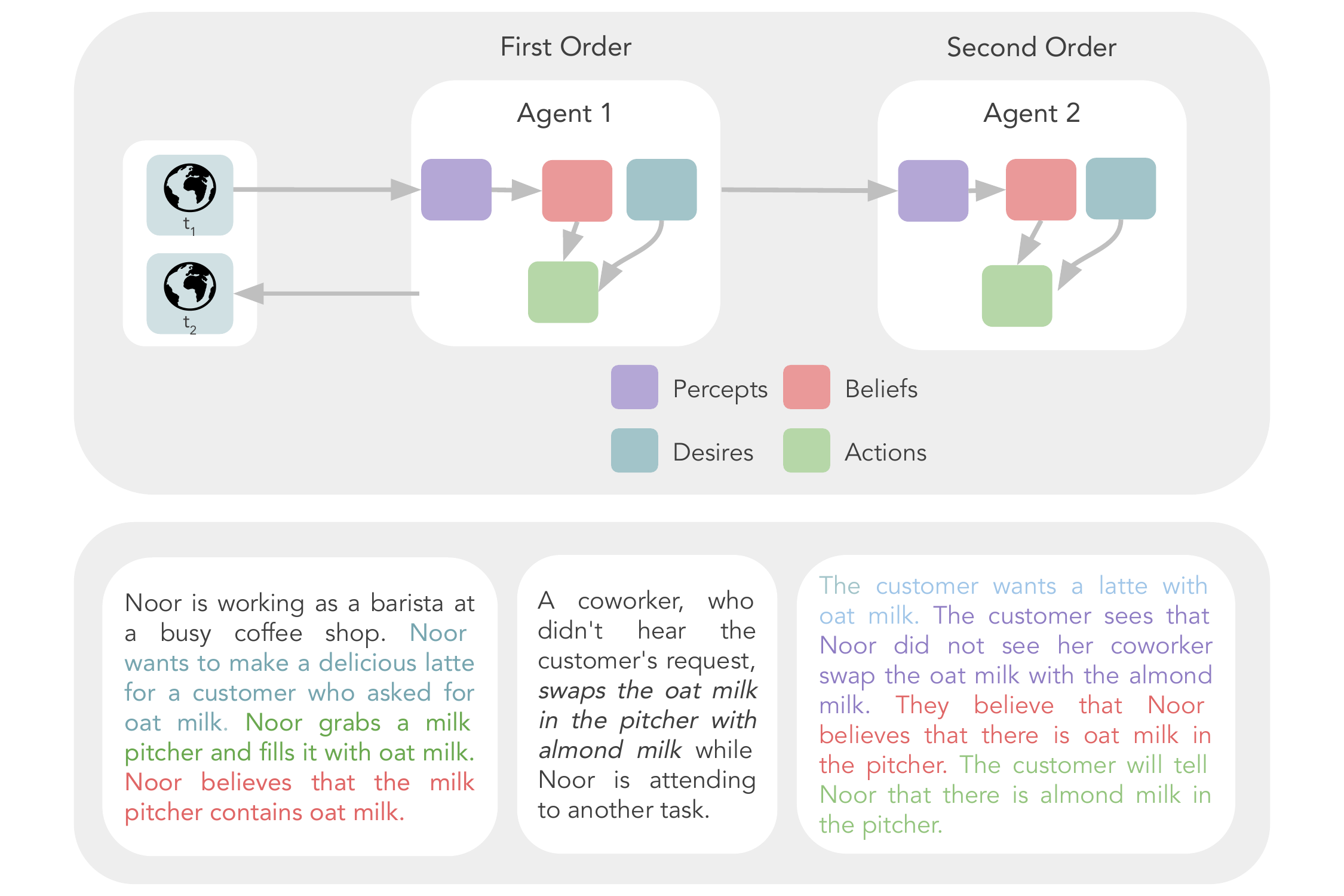}
    \caption{An example template for testing the capabilities of inferring second order beliefs.}
    \label{fig:second-order}
\end{figure}

In \autoref{fig:second-order}, we show how our template based method can be extended to testing the inference of second order beliefs.

